\documentclass{article}
\PassOptionsToPackage{numbers, compress}{natbib}
\usepackage[final]{neurips_2022}
\usepackage[utf8]{inputenc} 
\usepackage[T1]{fontenc}    
\usepackage[colorlinks=true,citecolor=red]{hyperref} 
\usepackage{url}            
\usepackage{booktabs}       
\usepackage{amsfonts}       
\usepackage{nicefrac}       
\usepackage{microtype}      
\usepackage{xcolor}         
\usepackage{color}
\usepackage{arydshln}
\usepackage{graphicx}
\usepackage{amsmath,bm}
\usepackage{amssymb}
\usepackage{array}
\usepackage{comment}
\usepackage{booktabs}
\usepackage{subfigure}
\usepackage{caption}
\usepackage{colortbl} 
\usepackage{multirow}
\usepackage{makecell}
\usepackage{wrapfig}

\usepackage[normalem]{ulem}
\usepackage{tikz}
\usepackage{algorithm}
\usepackage{algorithmic}

\definecolor{mypurplex}{RGB}{173,173,173}

\title{LasUIE: Unifying Information Extraction with Latent Adaptive Structure-aware Generative Language Model}

\author{
    Hao Fei\,$^{1}$ \quad Shengqiong Wu\,$^{1}$  \quad Jingye Li\,$^2$ \quad Bobo Li\,$^2$ \quad Fei Li\,$^2$\\
    \textbf{Libo Qin\,$^{1}$ \quad Meishan Zhang\,$^3$\thanks{Corresponding Author: Meishan Zhang} \quad Min Zhang\,$^3$ \quad Tat-Seng Chua\,$^{1}$} \\
    $^1$\,Sea-NExT Joint Lab, School of Computing, National University of Singapore \\ 
    $^2$Wuhan University \;\;\;\;\;\;\;  $^3$\,Harbin Institute of Technology (Shenzhen)\\
    \texttt{\{haofei37, liboqin, dcscts\}@nus.edu.sg} \quad \texttt{swu@u.nus.edu} \\
    \texttt{\{theodorelee, boboli, lifei\_csnlp\}@whu.edu.cn}\\
     \texttt{mason.zms@gmail.com} \quad \texttt{zhangmin2021@hit.edu.cn} \\
}

\begin{document}

\maketitle

\begin{abstract}
Universally modeling all typical information extraction tasks (UIE) with one generative language model (GLM) has revealed great potential by the latest study, 
where various IE predictions are unified into a linearized hierarchical expression under a GLM.
Syntactic structure information, a type of effective feature which has been extensively utilized in IE community, should also be beneficial to UIE.
In this work, we propose a novel structure-aware GLM, fully unleashing the power of syntactic knowledge for UIE.
A heterogeneous structure inductor is explored to unsupervisedly induce rich heterogeneous structural representations by post-training an existing GLM.
In particular, a structural broadcaster is devised to compact various latent trees into explicit high-order forests, helping to guide a better generation during decoding.
We finally introduce a task-oriented structure fine-tuning mechanism, 
further adjusting the learned structures to most coincide with the end-task's need.
Over 12 IE benchmarks across 7 tasks our system shows significant improvements over the baseline UIE system.
Further in-depth analyses show that our GLM learns rich task-adaptive structural bias that greatly resolves the UIE crux, the \emph{long-range dependence issue} and \emph{boundary identifying}.
Source codes are open at \url{https://github.com/ChocoWu/LasUIE}.
\end{abstract}

\section{Introduction}
\label{Introduction}

\begin{figure}[!t]
\centering
\includegraphics[width=1.0\columnwidth]{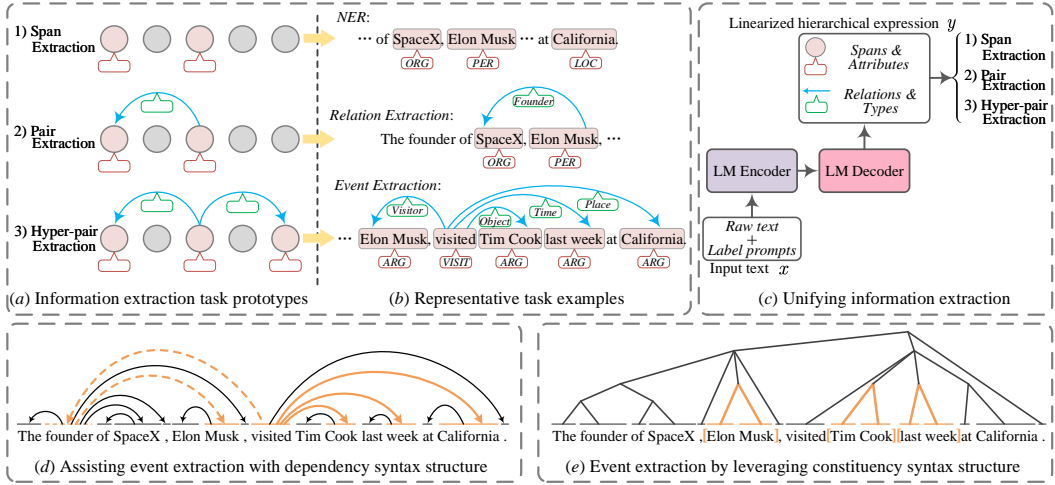}
\caption{
We reduce all the IE tasks into three prototypes (a) with representative examples (b).
We unify all IEs with an encoder-decoder GLM (c).
Both syntactic \emph{dependency} (d) and \emph{constituency} structure (e) plays a key but distinct role in IE, where the former helps solve \emph{long-range dependence} problem and the latter benefits \emph{boundary detection} issue.
Best viewed with zooming in.
}
\label{intro}
\vspace{-18pt}
\end{figure}

Information extraction (IE) is widely considered as one of the most kernel topics in natural language processing (NLP), which is defined as identifying the desired structural information from the unstructured texts \cite{andersen-etal-1992-automatic,surdeanu-etal-2003-using,maslennikov-chua-2007-multi,ma-hovy-2016-end,lin-etal-2018-adaptive,kolluru-etal-2020-imojie,0001JLLRL21}.
There is a variety of IE and IE-derived tasks, yet all of which revolves around predicting two key elements: \emph{mention spans} or/and \emph{their semantic relations}. 
For example as in Fig. \ref{intro}(b), NER detects the mention spans,
while RE recognizes each possible mention and its associated mention with relation.
In this regard, all the existing IE jobs can be reduced into three prototypes:  {span extraction}, {pair extraction} and {hyper-pair extraction}, as depicted in Fig. \ref{intro}(a).

In the era of deep learning, IE witnesses extraordinary developments, where especially the recent triumph of pre-trained language models (LMs) helps push the state-of-the-art (SoTA) IE performances amazingly \cite{devlin2019bert,alt-etal-2019-fine,JoshiCLWZL20,yan-etal-2021-unified-generative,wang-etal-2021-cleve}.
Prior related works mostly design particular models for certain IE tasks in isolation; while the latest SoTA progress \cite{UIEACL22} is achieved by unifying all IE tasks with a single encoder-decoder GLM, i.e., UIE.
As different IE tasks essentially share the similar nature (i.e., modeling span and relation features), it is proven that universally modeling multiple IE tasks helps further learning of general sharable knowledge from varying task sources, which makes UIE great potentials in real-world scenarios.
In this work we inherit this wisdom and also focus on UIE.

On the other hand, previous IE research extensively employs the external syntactic structure information, such as the dependency tree, for task improvements \cite{angeli-etal-2015-leveraging,miwa-bansal-2016-end,marcheggiani-titov-2017-encoding,jie-lu-2019-dependency,pouran-ben-veyseh-etal-2020-graph,0001WRZ22}.
Behind the enhancements is that IE structure corresponds much with the syntax structure explicitly, where the latter can essentially offer low-level linguistic bias for better learning the high-level semantic structure.
As exemplified in Fig. \ref{intro}(d), the dependency tree coincides much with structure of EE task as in Fig. \ref{intro}(b).
Importantly, some findings reveal that the LMs, being pre-trained on large corpus, capture structural syntax knowledge \cite{vig-belinkov-2019-analyzing,Assessing190105287,jawahar2019}, which gives rise to LMs' distinguishing promotion on IE.
Yet probing tasks show that the auto-learned structure representations is weak, which inevitably limits the LM efficacy for IE \cite{Probing2018,kovaleva-etal-2019-revealing,Probing19}.
Correspondingly, a line of researches fuse external syntax trees into LMs to reinforce the structure awareness, i.e., structure-aware LMs \cite{wang-etal-2019-tree,ahmed-etal-2019-need,li-etal-2021-improving-bert,bai-etal-2021-syntax}.

\textbf{Motivations.}
After carefully revisiting the existing literatures, we summarize four key limitations of syntactic structure-aware LMs that hamper IE from further improvements.
\textbf{First}, existing structure-aware LMs are mostly designed for one certain IE task (e.g., NER \cite{wang-lu-2020-two}, RE \cite{li-etal-2020-unified}) instead of UIE, leaving the shared IE knowledge and the task-invariant syntax features unexploited.
\textbf{Second}, current structure-aware LMs merely consider making use of one standalone type of syntax structures, i.e., mostly using the dependency trees \cite{miwa-bansal-2016-end,jie-lu-2019-dependency}.
We however argue that as one core grammar, constituency syntax can serve complementary contributions for IEs.
There are two common challenges of IEs: \emph{long-range dependence problem} and \emph{boundary identifying}, in which the dependency structure especially helps solve the former one \cite{miwa-bansal-2016-end,marcheggiani-titov-2017-encoding,pouran-ben-veyseh-etal-2020-graph} and the constituency syntax could mostly benefits the latter \cite{xu-etal-2015-lexicalized,marcheggiani-titov-2020-graph,quan2022compound}, as in Fig. \ref{intro}(d)\&(e).
Thus it is best to simultaneously model both two heterogeneous structures \cite{lepori-etal-2020-representations,fei-etal-2021-better}.
\textbf{Third}, existing works mostly integrate supervised syntax parse trees,
where unfortunately, either the amount of manually annotated syntactic data (e.g., PTB) is largely limited,
or the annotation noises from third-party parsers are inevitably introduced due to such explicit injection.
\textbf{Fourth}, parsing syntax comes with task-irrelevant or indirect substructures (e.g., in Fig. \ref{intro}(d) the black and dotted lines respectively), which would deteriorate the efficacy. 
Meanwhile, different IE tasks largely demand distinct structural feature biases, while current structure-aware LMs fail to fine-tune the structure knowledge to allow the structure bias best accord with end task's need.

\begin{figure}[!t]
\centering
\includegraphics[width=1.0\columnwidth]{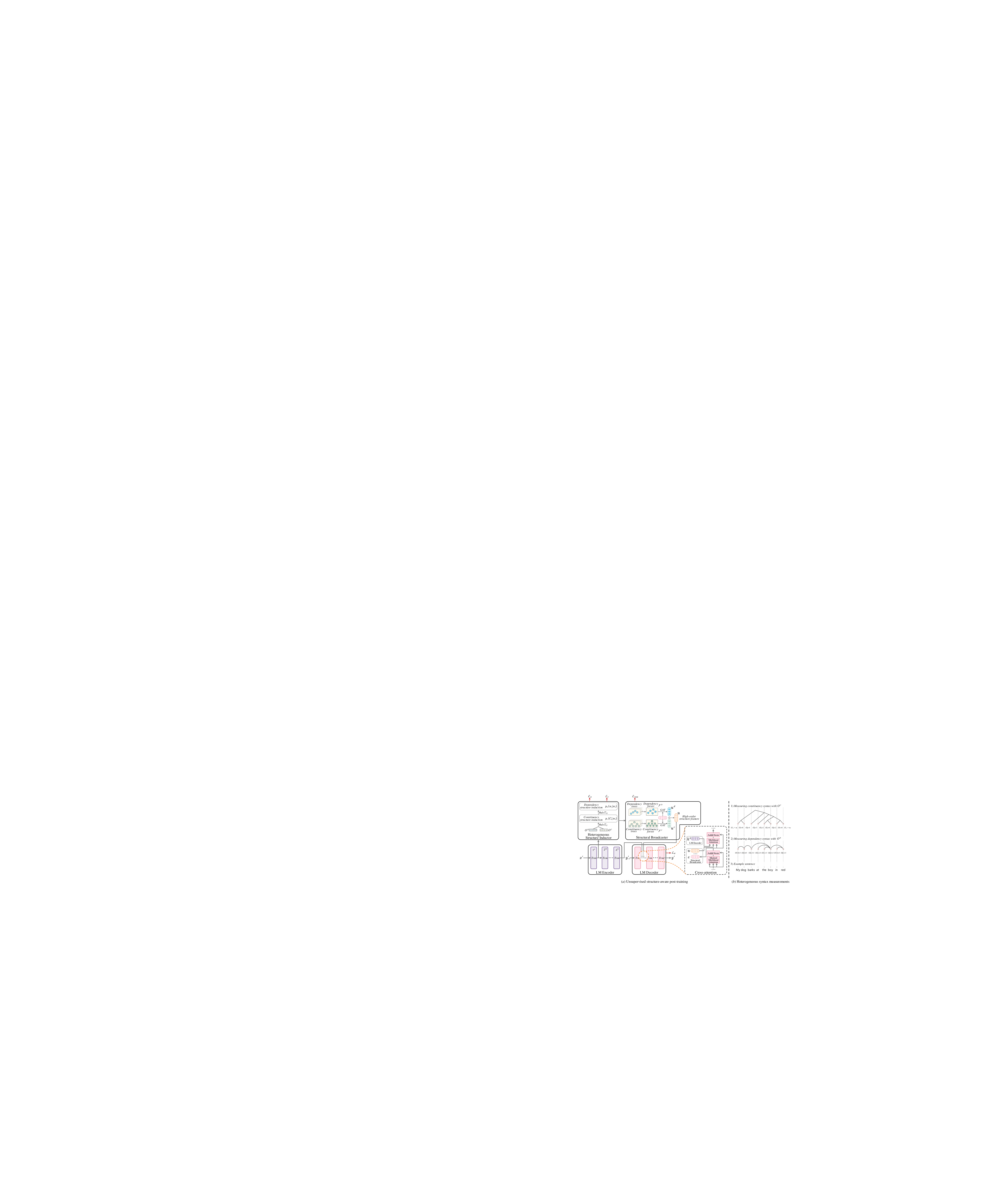}
\caption{
Our LasUIE framework under ($a$) unsupervised structure-aware post-training ($y^{'}_{\mapsto}$ refers to prediction $y^{'}$ with shift right in seq-to-seq procedure).
Heterogeneous structure inductor module generates both constituency and dependency structures via ($b$) two heterogeneous syntax measurements.
}
\label{framework1}
\vspace{-1.3em}
\end{figure}

\textbf{Contributions.}
On the above basis, we propose learning a \underline{l}atent \underline{a}daptive \underline{s}tructure-aware generative language model for \underline{UIE} (namely LasUIE).
First of all, we reduce UIE into three uniform prototypes, upon which we transform the UIE into generative paradigm with an encoder-decoder GLM, predicting the linearized hierarchical expression (i.e., spans\&attributes, relations\&types, as shown in Fig. \ref{intro}(c)).
Then, we adopt a three-stage of LM training procedure, where an additional structure-aware post-training is added between the pre-training and fine-tuning stages for structure learning.
Inspired by the progress of unsupervised grammar induction \cite{ShenLHC18,ShenTSC19,kim-etal-2019-compound,shen-etal-2021-structformer}, we design a heterogeneous structure inductor (HSI) module, where two heterogeneous syntactic structures are simultaneously measured and automatically learned.
With HSI, our GLM initialized with existing pre-trained parameters,
during post-training, performs unsupervised syntax induction based on unlabeled texts without relying on external syntax parses or any annotation labor (cf. Fig. \ref{framework1}).

Since the induced latent structural representations may be squeezed aside by the mainstay contextual representations in LM encoder, we further enhance the utility of syntax by introducing a structural broadcaster (SB) module (cf. Fig.\ref{framework1}).
SB compacts multiple varying latent trees from different encoding attention heads into an explicit constituency-like and a dependency-like forest respectively.
During each decoding step, two heterogeneous syntactic forests are utilized to produce high-order features at global level for guiding better content generation.
Finally, during the prompt-based fine-tuning stage we perform task-oriented structure adaptive tuning to narrow the gaps between the induced syntactic and task-specific structures (cf. Fig. \ref{framework2}).
With policy gradient we dynamically adjust the attributes of two heterogeneous structures according to the feedback of end task performance.

Extensive experiments are performed on 12 representative data across 7 IE tasks.
On both the supervised and low-resource settings our framework consistently shows improvements over the baseline systems.
Via further analyses we verify that
\textbf{1)} unifying IE tasks by further modeling structure information in LM benefits IE substantially, especially in the low-resource scenario.
\textbf{2)} Integrating two heterogeneous structures brings mutual advantages for UIE, helping fully resolve the boundary identifying and long-range dependence issue.
\textbf{3)} Automatically inducing latent structures in LM with further task-oriented structural adaptation learning significantly consolidates the efficacy of structure knowledge for end tasks.
\textbf{4)} Different types of IE tasks rely subtly on varying structural bias, all of which can be flexibly learned and correctly satisfied by our system.

\vspace{-6pt}
\section{Related Work}
\label{Related Work}

\vspace{-3pt}
IE is a long-standing research topic in NLP, which includes various tasks as well as growing derivations \cite{andersen-etal-1992-automatic,surdeanu-etal-2003-using,maslennikov-chua-2007-multi,ma-hovy-2016-end,W2NER22,ORL22}.
We reveal that essentially all the IE tasks can be summarized into three main prototypes, according to the combination numbers of `mention span' and `semantic relation' prediction targets:
1) \textbf{span extraction}, e.g., named entity recognition (NER) \cite{cucerzan-yarowsky-1999-language}, aspect-based sentiment analysis (ABSA) \cite{tang-etal-2016-effective}, aspect-term extraction (ATE) \cite{li-lam-2017-deep};
2) \textbf{pair extraction}, e.g., relation extraction (RE) \cite{zelenko-etal-2002-kernel,li-etal-2021-mrn}, aspect-opinion pair extraction (AOP) \cite{ZhaoHZLX20}, aspect-based sentiment triplet extraction (ASTE) \cite{PengXBHLS20};
and 3) \textbf{hyper-pair extraction}, e.g., event extraction (EE) \cite{halpin-moore-2006-event}, semantic role labeling (SRL) \cite{gildea-jurafsky-2000-automatic}, opinion role labeling (ORL) \cite{kim2006extracting,shi-etal-2022-effective}.
Mostly prior IE researches all solve one particular task exclusively (or one specific IE type) \cite{miwa-bansal-2016-end,wang2020tplinker,lin-etal-2020-joint,Wu0RJL21,zheng-etal-2021-prgc}, while they may unfortunately ignore certain task-invariant universal IE features.
In this work, we consider the line of UIE, unifying all IE tasks to exploit the shared IE knowledge.
And based on the above UIE prototypes, we develop a LM-based unified framework with generative paradigm.

Many efforts are paid for building LMs to handle IE tasks by taking advantages of the knowledge from large-scale pre-training \cite{devlin2019bert,alt-etal-2019-fine,JoshiCLWZL20,yan-etal-2021-unified-generative,wang-etal-2021-cleve}.
Another line of IE researches propose injecting external knowledge into LMs or GLMs, such as knowledge graph (KG) \cite{liu-etal-2019-knowledge-augmented,kawintiranon-singh-2021-knowledge,zhang-etal-2021-smedbert,FeiRZJL21}, syntax structure information \cite{wang-etal-2019-tree,ahmed-etal-2019-need,li-etal-2021-improving-bert,bai-etal-2021-syntax}.
Comparing to the integration of domain-specific KG information for certain IE tasks, syntactic information would provide much broader generic features in the scope of UIE.
The very latest research attention of LMs has been focused on the GLMs, the encoder-decoder paradigm LMs.
GLMs transform various NLP tasks into a unified seq-to-seq scheme with some properly-designed prompt texts as additional inputs \cite{lewis-etal-2020-bart,RaffelSRLNMZLL20,yan-etal-2021-unified-generative}.
Very recently, Lu et al. (2022) \cite{UIEACL22} pioneer the UIE by casting the IE structure prediction into text generation with a GLM, with which our UIE modeling shares the same spirit.
We however note that our work can advance in two major aspects.
First, we consider the integration of additional structural knowledge in GLMs for UIE enhancements.
Besides, \cite{UIEACL22} require supervisedly pre-training their UIE GLM on a large-scale annotated IE corpus, while our system automatically induces structure knowledge based merely on unlabeled texts without any further annotation and labor.

This work also closely relates to the line of structure-aware LMs.
On the one hand, some researches propose directly introducing external syntax trees into LMs to reinforce the structure awareness.
They mostly take the Transformer-based LMs as backbone, and fuse the syntax signals (annotations) from external parsers or PTB corpus by modifying the Transformer attentions \cite{wang-etal-2019-tree,li-etal-2021-improving-bert,bai-etal-2021-syntax}.
Another line of structure-aware LMs directly induce syntax structure into LMs automatically, a.k.a., unsupervised grammar induction \cite{ChoiYL18,WilliamsDB18,ShenLHC18,drozdov-etal-2019-unsupervised-latent,ShenTSC19,kim-etal-2019-compound}.
We in this work borrow the success of unsupervised grammar induction, inducing rich structure information for LMs for better UIE.
Inspired by the foundation of syntax distance measurements \cite{ShenLHC18,fei-etal-2020-retrofitting,shen-etal-2021-structformer}, we propose to induce linguistic structures and compose both the constituency and dependency syntax structures simultaneously.

\vspace{-8pt}
\section{Unifying Information Extraction with Text Generative Paradigm}
\label{Task Modeling}

\vspace{-4pt}
As aforementioned, we reduce all IE jobs into the predictions of few structural elements: 1) \emph{spans} and 2) \emph{relations}, and without losing the generality of UIE, we also consider the predictions of 3) \emph{span attributes} and 4) \emph{relation types}.
Using different combinations of the structural elements properly can construct a hierarchical IE structure.
In other words, we can arrange these elements into a sequential textual expression, from which the actual IE target can be easily restored.
Based on this we unifiedly model all IE tasks (i.e., UIE) by transforming the structure prediction into text generation.
Fig. \ref{intro}(c) illustrates the main idea.
Such generative scheme also enables to take the advantage of the recent achievement of GLMs, such as BART \cite{lewis-etal-2020-bart} and T5 \cite{RaffelSRLNMZLL20}.
Taking an existing GLM as backbone, we reach the goal of end-to-end UIE as well as complex IE, such as overlapped and discontinuous cases \cite{zeng-etal-2018-extracting,fei2020boundaries,li-etal-2021-span}.
Our LM encoder takes input text (i.e., $x$), and the decoder produces the linearized hierarchical expression (LHE), i.e., $y$.
Fig. \ref{framework2} illustrates the input and output implications.

\textbf{Input.}
The input text $x=\{w_1,\cdots,w_n\}$ includes the raw sentence and task-specific label prompts.
The label prompts contain the pre-defined task-specific labels, including \emph{span attributes} (`{\small \emph{Attr}}') and \emph{relational types} (`{\small \emph{Type}}'), where each label is separated by a `{\small <SPAN>}' or a `{\small <REL>}' marker.
Parts of the `{\small \emph{Attr}}' and `{\small \emph{Type}}' labels will be copied and output in $y$.
We also insert a task identifier token `{\small <TASK>}' to inform the model which task to predict.

\textbf{Output.}
The output text $y$ is a linearized hierarchical expression that describes how the structural elements organize into the target structure, as depicted in Fig. \ref{framework2}.
For example, in span extraction, $y$ should be a list of text spans and attribute labels, i.e., `{\small $\{(Span, Attr),\cdots\}$}'.
In pair extraction $y$ is a list of pairs, where a pair is represented as `{\small $(Span_i, Attr_i \, [Type_k] (Span_j, Attr_j))$}' in which {\small $Span_j$} is a subordinate mention of {\small $Span_i$} with a semantic relation {\small $Type_k$}.
For hyper-pair extraction $y$ is a list of hyper-pairs represented as `{\small $(Span_i, Attr_i \, [Type_k] (Span_j, Attr_j) \, [Type_m] (Span_k, Attr_k) \cdots)$}'.

It is also noteworthy that our LHE takes a similar scheme with \cite{UIEACL22}, but with differences.
For example, in our scheme all the mention comes with an associated attribute label in any IE prototype; while in \cite{UIEACL22} the subordinate mentions have no attribute labels.
Thus, our design could be more generalizable.

\vspace{-8pt}
\section{Learning Latent Adaptive Structure-aware Generative Language Model}
\label{Latent Adaptive Structure-aware Language Model}

\vspace{-3pt}
\subsection{Overall Framework}
\label{Overall Framework}

\vspace{-3pt}
The overall framework is built upon a Transformer-based encoder-decoder GLM, based on which we additionally add 1) a heterogeneous structure inductor module at top of the encoder for structural learning, 2) a structural broadcaster module between GLM encoder and decoder for enhancing the structural feature utility.
Fig. \ref{framework1} shows the overall architecture of our LasUIE GLM.

LasUIE takes a three-stage training process, where a {structure-aware post-training} is inserted between the pre-training and fine-tuning stages for structure learning. 
LasUIE takes an existing well pre-trained GLM parameters (e.g., BART, T5) as initiation.
During structure-aware post-training stage
our GLM carries out unsupervised syntax induction based on unlabeled plain texts (cf. $\S$\ref{Unsupervised Structure-aware Post-training}).
Thereafter, LasUIE is fine-tuned on the in-house training data, along with which  we perform task-oriented structure adaptive tuning (cf. $\S$\ref{Task-oriented Structure Fine-tuning}).
We also note that LasUIE takes a consistent paradigm of text-to-text generation throughout the whole three stages, which ensures a minimum information loss from the early trainings to the final predicting.

\vspace{-5pt}
\subsection{Unsupervised Structure-aware Post-training}
\label{Unsupervised Structure-aware Post-training}

\vspace{-3pt}
\textbf{Heterogeneous structure inductor.}
As cast earlier, although LMs are able to learn certain linguistic knowledge from generic pre-training, the signal strength of learned syntax is quite weak to contribute IE enough \cite{Probing19,kovaleva-etal-2019-revealing}.
In the structure-aware post-training stage, we aim to unsupervisedly enrich our GLM with sufficient structural knowledge, reinforcing the awareness of linguistic syntax.

Inspired by Shen et al. (2021) \cite{shen-etal-2021-structformer}, we explore a heterogeneous structure inductor (HSI) stacked on top of GLM encoder to reach the above goal.
HSI induces linguistic structures based on the foundation of syntax distance measurements \cite{ShenLHC18}.
We employ two heterogeneous syntax measurements, i.e., $O^C$=$\{o^c_1,\cdots,o^c_{n-1}\}$ ($o^c_{<1}$=$o^c_{>n-1}$=$\infty$) for measuring constituency syntax, and $O^D$=$\{o^d_1,\cdots,o^d_{n}\}$ for measuring dependency syntax.
As illustrated in Fig. \ref{framework1}(b), $o^c_i$ is a real value depicting the height of the lowest common ancestor between two consecutive words $w_i$ and $w_{i+1}$; while $o^d_i$ is a real value describing the spanning distance between the words linking to $w_i$.
Intuitively, bigger $o^c_i$ means bigger information divergence of the split point between the two sides of phrasal span, and larger $o^d_i$ implies wider range of connections, i.e., longer-term dependent relations.
As revealed that the syntax features are best learned at lower layer of GLM encoder \cite{jawahar2019}, HSI thus takes the first-layer encoding representations $\bm{h}_i^1$ as input and produce syntax context representations via convolution operation: $\bm{h}^*_i$=Conv($\bm{h}_i^1$).
Based on $\bm{h}^*_i$, HSI represents $o^c_i$ and $o^d_i$ as:
\begin{equation}\small\label{two-measurements}
o^c_i = \bm{V}^c \text{Tanh}( \bm{W}[\bm{h}^*_i;\bm{h}^*_{i+1} ])\,, \quad
o^d_i = \bm{V}^d \text{Tanh}( \bm{W}\bm{h}^*_i ) \,.
\end{equation}

Then, two rules are made for generating two heterogeneous syntax based on the two measurements \cite{shen-etal-2021-structformer}, which also helps coordinate two types of structures so that they can co-exist together and legally.

\setlength\parindent{14pt}

\indent $\blacktriangleright$ \textbf{Rule $\Gamma_{C}$}:
\emph{A smallest constituent span $C_{[l,r]}$ of $w_i$ ($l$$<$$i$$<$$r$) should satisfy ($o^c_{l-1}$$>$$o^d_i$)$\&$($o^c_{r}$$>$$o^d_i$).}
For example as in Fig. \ref{framework1}(b), $o^c_{2}$(=4)$>$$o^d_3$(=3.5) and $o^c_{8}$(=$\infty$)$>$$o^d_3$, thus $C_{[3,8]}$ is the valid minimum span for $w_3$.

\indent $\blacktriangleright$ \textbf{Rule $\Gamma_{D}$}:
\emph{Generalizing $w_i$ as a potential span $C_{[l=i,r=i]}$, the dependent head of any word in $C_{[l,r]}$ is $w_j\leftarrow\mathop{\text{argmax}}_{k\in [l,r]}(o^d_k)$.
}
For example, the maximum $o^d$ in constituent span of $C_{[3,8]}$ is $o^d_6$=4.5, thus dependent head of the word in $C_{[3,8]}$ is $w_6$.

\setlength\parindent{0pt}

Based on rule $\Gamma_{C}$ we first generate all possible phrasal spans and organize them into a constituency tree $\mathcal{T}^C$, then constructing the dependency tree $\mathcal{T}^D$ according to rule $\Gamma_{D}$.
We parameterize the above structure construction process so as to make it all differentiable, i.e., by describing into the probabilistic perspective.
We represent the span $C_{[l,r]}$ distribution as:
\begin{equation}\small
\setlength\abovedisplayskip{2pt}
\setlength\belowdisplayskip{2pt}
\begin{aligned}\label{eqn1}
p_c(c_k|w_i) &= p(w_l|w_i)\cdot p(w_r|w_i) \\
     &= [\sigma(o^d_i - \mathop{\text{Max}}_{ k\in [l,i)}(o^c_k)  ) -  \sigma(o^d_i - \mathop{\text{Max}}_{k\in (l,i)}(o^c_k)  )] \cdot
     [\sigma(o^d_i - \mathop{\text{Max}}_{k\in [i,r)}(o^c_k)  ) -  \sigma(o^d_i - \mathop{\text{Max}}_{k\in [i,r]}(o^c_k)  )] ,
\end{aligned}
\end{equation}
where $c_k$ is a short hand for $C_{[l,r]}$, $\sigma$ is a sigmoid function.
We then depict the rule $\Gamma_{D}$, and represent the word-word dependent distribution:
\setlength\abovedisplayskip{2pt}
\setlength\belowdisplayskip{2pt}
\begin{equation}\small\label{eqn2}
 p_d(w_j|w_i) = p_d(w_j|c_k) \cdot p_c(c_k|w_i) = p_c(c_k|w_i) \cdot \exp(\bm{h}^L_j) /  \sum^r_{k=l}   \exp(\bm{h}^L_k)  \,,
\end{equation}
where we use the top-layer encoder representation $\bm{h}^L_i$, by which we encourage the final encoding representations to learn from the low-layer  syntax-rich representations.
We note that the above structure induction is carried out in each of multi-head attention blocks (total $M$) in Transformer.
This means that multiple distinct syntax trees of each type (i.e., $\mathcal{T}^C$ and $\mathcal{T}^D$) will be induced.

\textbf{Structural broadcaster.}
It is a high chance that in GLM encoder the mainstay contextual representations will weaken the structural features and thus hurt the structure utility at decoder.
To combat this, we propose a SB module, by which we explicitly collect varying trees of a type and compacted them into a forest, respectively, i.e., constituency forest $\mathcal{F}^C$ and dependency forest $\mathcal{F}^D$.
According to prior studies \cite{mi-huang-2008-forest,ma-etal-2018-forest,song-etal-2019-leveraging},
comparing to the optimal 1-best syntax tree, a compact forest advances in higher structure recall, which allows to learn a better bias for task.
In SB, the structural priors from the syntax forests are explicitly broadcast into each decoding step for guiding the generation process.

Technically, SB selects and ranks candidate tree substructures based on the probabilistic confidence (Eq.\ref{eqn1}\&\ref{eqn2}), which are then compacted into a forest based on the K-best maximum spanning tree (MST) algorithm  \cite{agic-2012-k,zmigrod-etal-2021-finding}.
We then model the two types of forests with a graph attention model (GAT) \cite{VelickovicCCRLB18} respectively,
during which the decoding representation $\bm{e}$ at each step is attended to spot the high-order structural feature $\bm{u}$ at global level:
\setlength\abovedisplayskip{2pt}
\setlength\belowdisplayskip{2pt}
\begin{equation}\small \label{GAT_1}
\bm{u}^{c/d} = \text{GAT}( \mathcal{F}^{C/D} , \bm{e}  \quad | \quad \bm{h}^{1} ) \,, 
\end{equation}
\begin{equation}\small \label{GAT_2}
\bm{u} = \bm{u}^{d} \oplus \bm{u}^{c} \,. 
\end{equation}
Further via a cross-attention operation (cf. Fig. \ref{framework2}) we navigate the encoder representation $\bm{h}^{L}$ and the structural feature $\bm{u}$ into the updated encoder representation $\bm{e}^{*}$:
\begin{equation}\small\label{cross-att}
\setlength\abovedisplayskip{2pt}
\setlength\belowdisplayskip{2pt}
 \bm{e}^{*} = \text{Softmax}( \frac{\bm{h}^{L} \cdot \bm{u}}{\sqrt{d}}  ) \cdot \bm{e}  \,.
\end{equation}

\normalsize

\textbf{Post-training objectives.}
The first objective is performing seq-to-seq style language modeling, i.e., `corrupting + reconstructing' the inputs \cite{lewis-etal-2020-bart}, which is identical to the pre-training objectives.
We denote the language modeling loss as $\mathcal{L}_W$.

Along with the language modeling we then promote the unsupervised structure induction, including the one for dependency syntax and the one for constituency syntax:
\begin{align}\label{Loss_syn}
\small  \mathcal{L}_D &= - \begin{matrix} \sum^{M}_m \sum_i^n \sum_j^n \end{matrix} \log p_d(w_j|w_i) \,, \\
\small \mathcal{L}_C &= - \begin{matrix} \sum^{M}_m \sum_i^n \sum_k^K \end{matrix} [ \log p_c(c_k|w_i) +  \log (\exp\phi(c_k) / \mathcal{Z}) ] \,,
\end{align}
where $\phi(c_k)$=$\frac{(\bm{h}^c)^T \cdot \bm{h}^L_{[l,r]} }{|| \bm{h}^c||\cdot || \bm{h}^L_{[l,r]} ||}$ is a span similarity score between the constituent phrase $c_k$ and the counterpart text span derived from the top-layer encoder.
$\mathcal{Z}$ is for normalization.

We also perform structure diversifying regularization (SDR), putting constraints on the varying trees induced from different multi-head encoder attentions so as to ensure structure diversification.
\begin{equation}\small\label{Loss_sdr}
 \mathcal{L}_{SDR} = - \begin{matrix} \sum^{M}_m \sum^{M}_k \end{matrix} || A_m \odot A_k || \,, \quad  k \neq m \,,
\end{equation}
where $A_m$ or $A_k$ is an attention map.
We put all the above objectives together as the post-training target: 
$\mathcal{L}_{PRT}=\mathcal{L}_W+\mathcal{L}_D+\mathcal{L}_C+\mathcal{L}_{SDR}$.

\subsection{Task-oriented Structure Fine-tuning}
\label{Task-oriented Structure Fine-tuning}

\begin{wrapfigure}{r}{9cm}
\vspace{-1.3em}
\begin{center}
\includegraphics[width=1\linewidth]{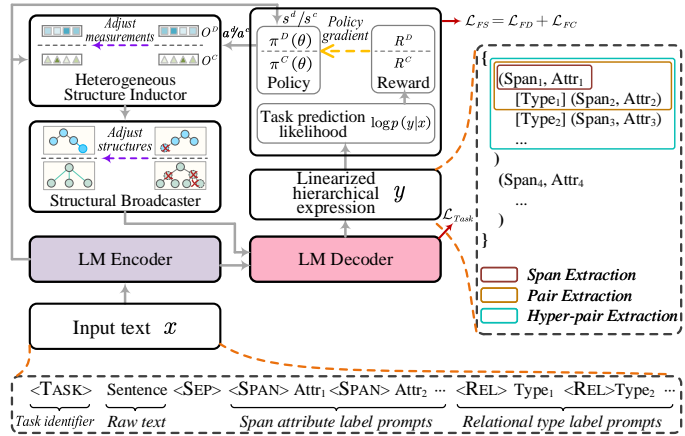}
\end{center}
\caption{
Fine-tuning our GLM with structure adaptive learning.
}
\label{framework2}
\end{wrapfigure}

After structure-aware post-training, our GLM is finally fine-tuned on a specific terminal IE task to learn the on-demand features.
This target is an empirical risk minimization with cross-entropy loss:
\begin{equation}\small\label{Loss_task}
 \mathcal{L}_{Task} = - \begin{matrix} \sum^{D} \end{matrix} \log p(y|x) \,,
\end{equation}
where $D$ is the mini-batch size.

\vspace{1pt}
Meanwhile, we perform further task-oriented structural fine-tuning, adapting the learned structure information to the task-specific IE structures,
for example, in dependency structure pruning those trivial word-word connections and adjusting the range of dependent paths;
in constituency structure refining the phrase widths and granularities.
Our main idea is to amend the syntax attribute (i.e., dependency links and constituent compositions) by directly taking the feedback of end task performance.
Therefore, we employ the stochastic policy gradient algorithm \cite{Williams92}.
As shown in Fig. \ref{framework2}, the actions $a_i^{c/d} \in (-1,1)$ are real values sampled with probabilities from a Gaussian distribution, by which we maintain a continuous control over syntax measurements, i.e., $O^{D}$ and $O^{C}$.
\setlength\abovedisplayskip{3pt}
\setlength\belowdisplayskip{3pt}
\begin{align}
\small   \bar{a}_i^{c/d} \sim \pi^{C/D}&(\bm{s}_i^{c/d};\theta^{c/d})=\mathcal{N}(0,I)  \,, \\
\small  a_i^{c/d} &= 2\sigma(\bar{a}_i^{c/d})-1    \,, \\
\small  o_i^{c/d} &:= a^{c/d} + o_i^{c/d}  \,,
\end{align}
where the $\bm{s}_i^{c/d}$=($\bm{h}_i^1\oplus\bm{h}_i^{*}\oplus\bm{h}_i^L$) are the state representations as the inputs of the policy agents.
The policy agents $\pi^{C}(\theta^{c})$ and $\pi^{D}(\theta^{d})$ are two parameterized two-layer feedforward networks, respectively.
We design the reward of the policy as the probability of correct task prediction, such that the structure adjustments are directly supervised by terminal task's signals: $R^{C/D}=\log p(y|x)$.
The learning target of each policy is to maximize the corresponding expected reward:
\setlength\abovedisplayskip{3pt}
\setlength\belowdisplayskip{3pt}
\begin{align}
\small   \mathcal{L}_{FD} &= -  \sum_{(a^d_1 s^d_1\cdots a^d_n s^d_n)} \prod_i  p(a^d_i|\bm{s}^d_i;\theta^d) \cdot R^D_i  \,, \\
\small  \mathcal{L}_{FC} &= -  \sum_{(a^c_1 s^c_1\cdots a^c_n s^c_n)} \prod_i  p(a^c_i|\bm{s}^c_i;\theta^c) \cdot R^C_i  \,.
\end{align}
We summarize all the fine-tuning targets:
$\mathcal{L}_{FT}$=$\mathcal{L}_{Task}+\mathcal{L}_{FS}$, where  $\mathcal{L}_{FS}$=$\mathcal{L}_{FD}+\mathcal{L}_{FC}$.

\vspace{-8pt}
\section{Experiments}
\label{Experiments}

\vspace{-3pt}
\subsection{Setups}
\label{Setups}

\vspace{-3pt}
We take the pre-trained T5 Base as the default backbone GLM.
We use the plain texts from \emph{Wikipedia}\footnote{\url{https://autonlp.ai/datasets/wikipedia-news-corpus}} and \emph{BooksCorpus}\footnote{\url{https://huggingface.co/datasets/bookcorpus}} corpora for the post-training.
To cover all three UIE prototypes, we consider 7 representative IE tasks with corresponding data:
1) NER: CoNLL03 \cite{tjong-kim-sang-de-meulder-2003-introduction}, OntoNote \cite{pradhan-etal-2013-towards}, ACE04 \cite{DoddingtonMPRSW04}, ACE05 \cite{ACE5};
2) RE: CoNLL04 \cite{roth-yih-2004-linear}, NYT \cite{RiedelYM10}, ACE05 \cite{ACE5};
3) AOP: Res14 \cite{pontiki-etal-2014-semeval};
4) ASTE: Res14 \cite{pontiki-etal-2014-semeval};
5) ORL: MPQA \cite{WiebeWC05};
6) SRL: CoNLL12 \cite{pradhan-etal-2013-towards};
7) EE: ACE05 \cite{ACE5}.
Each dataset has its own split, and we follow the same practice of the relevant prior works when using it.

We verify the IE performances under the traditional separate scheme and the recent unified scheme, respectively.
\textbf{\emph{1) In separate IE}}, we compare with the current SoTA systems (all using Large version LM/GLM) of each specific data; meanwhile we implement a T5 (Base version) system, namely GEN-T5, using the same generative manner (based on prompt input, generating LHE as ours) for running each task individually.
We also retrofit the GEN-T5 system by injecting into the external syntax parse trees via additional training on the syntax annotated corpus, including the dependency syntax (\emph{+DepSyn}), constituency syntax (\emph{+ConSyn}) and both two types syntax (\emph{+Dep\&ConSyn}), respectively.
\textbf{\emph{2) In unified IE}}, we mainly make comparisons with the current UIE system \cite{UIEACL22}.
Note that the default UIE model (marked as UIE$^{*\dag}$) in raw paper uses T5 Large and meanwhile takes additional supervised pre-training on the large-scale
IE corpus (the version without supervised IE pre-training marked as UIE$^{*}$). 
To ensure fair comparisons, we re-implement their system with T5-Base parameters and without supervised IE pre-training, marked as UIE.
Same as to GEN-T5, we also retrofit the UIE model by integrating heterogeneous syntax parse trees in different combinations.

Following each of previous works, we use the F1 evaluation metrics.
For each task, we consider the end-to-end prediction.
For example, for the span extraction (NER), we measure if both the mention span and the mention attribute are correct.
For the pair(/hyper-pair) extraction, we measure if the span boundary \& span attribute \& relation \& type are all correct simultaneously.

\vspace{-3pt}
\subsection{Main Results}

\vspace{-3pt}
We present the overall comparison results on various IE tasks in Table \ref{main-1} and Table \ref{main-2} under the fully-supervised and low-resource scenarios, respectively.
As can be seen, 
our proposed LasUIE framework consistently outperforms the baseline UIE and other SoTA models on all tasks in both two learning scenarios, under both the Large or Base T5 initiations.
This demonstrates the efficacy of our proposal.
Also we compare the counterparts between M2-M5 and M9-M12, where the only difference between these generative methods lies in the separate or unified modeling of IE.
From the results we learn that the unified modeling of IE (i.e., UIE) is more effective than the traditional separate modeling of specific IE task.
This verifies that the universal modeling helps share the task-invariant IE features, coinciding with the findings in \cite{UIEACL22}.

\begin{table*}[!t]
\fontsize{8}{9.2}\selectfont
 \setlength{\tabcolsep}{0.2mm}
  \caption{
Overall IE performances by different methods (all using LM/GLM).
Models with $*$ (M1, M6, M7 \& M8) refers to the use of Large version LM, where scores by M1, M6 \& M7 are copied from their raw paper \cite{UIEACL22}.
UIE$^{*\dag}$ (M6) takes additional supervised pre-training on the large-scale IE corpus.
\textbf{Bold}: the best results among the comparisons using Large and Base LMs, respectively.
}
\begin{center}
\resizebox{1\columnwidth}{!}{
  \begin{tabular}{ccl|ccccc|cccccc|cccc|c}
\toprule
 \multicolumn{3}{c|}{\multirow{3}{*}{ Task\&Data}}&   \multicolumn{4}{c}{\bf\em Span Extraction} &\phantom{} &  \multicolumn{5}{c}{\bf\em Pair Extraction} &\phantom{}& \multicolumn{3}{c}{\bf\em Hyper-pair Extraction} &\phantom{} & \multicolumn{1}{c}{\multirow{3}{*}{\em Avg.}} \\
\cmidrule(r){4-8}\cmidrule(r){9-14}\cmidrule(r){15-18}
&&  &   \multicolumn{4}{c}{\scriptsize \bf NER} &\phantom{} & \multicolumn{3}{c}{\scriptsize \bf RE}  & \multicolumn{1}{l}{\scriptsize \bf AOP} & \multicolumn{1}{l}{\scriptsize \bf ASTE} &\phantom{} &  \multicolumn{1}{l}{\scriptsize \bf ORL} & \multicolumn{1}{c}{\scriptsize \bf SRL} & \multicolumn{1}{c}{\scriptsize \bf EE} &\phantom{} & \\
\cmidrule(r){4-8}\cmidrule(r){9-11}\cmidrule(r){12-12}\cmidrule(r){13-13}\cmidrule(r){15-15}\cmidrule(r){16-16} \cmidrule(r){17-17}
&&  &     {\scriptsize CoNLL03} & {\scriptsize OntoNote} &  {\scriptsize ACE04} & {\scriptsize ACE05} &\phantom{} &   {\scriptsize CoNLL04} &   {\scriptsize NYT} &   {\scriptsize ACE05} &   {\scriptsize Res14} &   {\scriptsize Res14} &\phantom{} &   {\scriptsize MPQA} &    {\scriptsize CoNLL12} &   {\scriptsize ACE05} &\phantom{} &\\
\midrule

\multicolumn{19}{l}{$\bullet$ \bf\em Separate IE  } \\
M1 &\phantom{} &  SoTA$^{*}$ &  93.2 &  91.9 &  86.8 &  84.7 &\phantom{} &  73.6 &  92.7 &  65.6 &  69.3 &  73.6 &\phantom{} &  53.0 &  73.5 &  48.3 &\phantom{} &  75.5 \\
\hline
M2 &\phantom{} &  GEN-T5 &  91.0 &  89.1 &  84.3 &  83.0 &\phantom{} &  69.4 &  90.3 &  60.2 &  62.5 &  71.8 &\phantom{} &  49.8 &  69.3 &  43.7 &\phantom{} &  72.0 \\
M3 &\phantom{} &   \quad\emph{+DepSyn} &   91.5 &  89.5 &  84.9 &  83.4 &\phantom{} &  70.3 &  91.8 &  62.4 &  64.3 &  72.6 &\phantom{} &  51.5 &  70.8 &  45.5 &\phantom{} &  73.2 \\
M4 &\phantom{} &   \quad\emph{+ConSyn} &   92.1 &  90.0 &  85.3 &  83.8 &\phantom{} &  69.8 &  90.9 &  61.5 &  63.1 &  72.3 &\phantom{} &  50.7 &  70.1 &  44.3 &\phantom{} &  72.8 \\
M5 &\phantom{} &   \quad\emph{+Dep\&ConSyn} &  92.3 &  90.4 &  85.3 &  84.0 &\phantom{} &  71.2 &  92.1 &  63.3 &  66.0 &  73.0 &\phantom{} &  51.8 &  71.3 &  46.2 &\phantom{} &  73.9 \\

\midrule\hline
\multicolumn{19}{l}{$\bullet$ \bf\em Unified IE }    \\ 
M6 &\phantom{} &  UIE$^{*\dag}$ &   93.0 &  / &   \bf 86.9 &  85.8 &\phantom{} &    75.0 &  / &   66.0 &  / &   74.5 &\phantom{} &  / &   / &   / &\phantom{} &   / \\
M7 &\phantom{} &  UIE$^{*}$ & 92.1 &  / &   86.5 &  85.5 &\phantom{} &  73.1 &  93.5 &  64.7 &  / &   / &\phantom{} &   / &   / &   / &\phantom{} &   / \\
M8 &\phantom{} &  \textbf{LasUIE$^{*}$} (Ours) &   \bf 93.2 &  \bf 93.0 &  86.8 &  \bf 86.0 &\phantom{} &  \bf 75.3 &  \bf 94.2 &  \bf 66.4 &  \bf   73.6 &    \bf 75.2 &\phantom{} &  \bf   57.8 &  \bf   76.3 &    \bf 51.7 &\phantom{} &    \bf 77.4 \\

\hline
M9 &\phantom{} &  UIE &   91.4 &  89.7 &  85.0 &  83.5 &\phantom{} &  70.5 &  91.0 &  61.6 &  65.8 &  72.8 &\phantom{} &  50.8 &  70.2 &  44.6 &\phantom{} &  73.1 \\
M10 &\phantom{} &    \quad\emph{+DepSyn} &   91.8 &  90.0 &  85.3 &  83.7 &\phantom{} &  71.2 &  92.0 &  62.9 &  67.6 &  73.5 &\phantom{} &  52.0 &  71.5 &  46.4 &\phantom{} &  74.0 \\
M11 &\phantom{} &    \quad\emph{+ConSyn} &   92.0 &  90.5 &  85.6 &  84.0 &\phantom{} &  70.8 &  91.3 &  62.1 &  66.1 &  73.1 &\phantom{} &  51.3 &  71.0 &  45.2 &\phantom{} &  73.6 \\
M12 &\phantom{} &    \quad\emph{+Dep\&ConSyn} &  92.3 &  90.7 &  85.8 &  84.5 &\phantom{} &  71.7 &  92.4 &  63.4 &  68.2 &  73.7 &\phantom{} &  53.6 &  72.6 &  47.0 &\phantom{} &  74.6 \\

\cdashline{1-19}
M13 &\phantom{} &   \textbf{LasUIE} (Ours) &   \bf   92.6 &    \bf 92.0 &  \bf   86.3 &  \bf   85.0 &\phantom{} &  \bf   73.2 &  \bf   93.0 &  \bf   64.4 &  \bf   70.2 &  \bf   74.8 &\phantom{} &  \bf   56.0 &  \bf   74.7 &  \bf   49.0 &\phantom{} &  \bf   75.9 \\

M14 &\phantom{} &    \quad w/o SB &   92.0 &  90.7 &  85.5 &  84.2 &\phantom{} &  71.5 &  91.8 &  62.9 &  68.3 &  73.4 &\phantom{} &  54.7 &  73.4 &  47.7 &\phantom{} &  74.6 \\
M15 &\phantom{} &    \quad w/o $\mathcal{L}_{SDR}$ &  92.2 &  91.6 &  86.2 &  84.8 &\phantom{} &  72.8 &  92.4 &  64.1 &  70.0 &  74.4 &\phantom{} &  55.5 &  74.0 &  48.6 &\phantom{} &  75.6 \\
M16 &\phantom{} &    \quad w/o $\mathcal{L}_{FS}$ &   92.4 &  91.4 &  85.9 &  84.7 &\phantom{} &  71.8 &  92.0 &  63.6 &  69.1 &  73.6 &\phantom{} &  54.2 &  73.0 &  47.1 &\phantom{} &  74.9 \\

\bottomrule
\end{tabular}
}
\end{center}
\vspace{-1.7em}
\label{main-1}
\end{table*}

\subsection{In-depth Analysis}

To aid better understanding the strengths of our method, we further present in-depth analyses from varying angles, i.e., by asking four key questions concentrating on the structure-aware GLM for UIE.

\paragraph{Q1: {Can fusing syntax structure knowledge into GLM contribute to UIE?}}

Let's compare the results in Table \ref{main-1}: M2 vs. M3\&M4\&M5 in separate IE setup, and M9 vs. M10\&M11\&M12 in unified IE setup, where \emph{either in separate or unified IE setup, integrating additional linguistic syntax features into GLM evidently improves all end task performances.}
Interestingly, different tasks can receive varying degree of improvements from the syntax features.
Importantly, we see in Table \ref{main-2} that with the aids of structure knowledge, the performances of low-resource transfer can be promoted, especially in the combination with the unified modeling of IE tasks.
This proves that \emph{the syntactic structures in GLM can serve as IE task-invariant features, further contributing to UIE.}

\begin{table*}[!t]
\fontsize{8}{10}\selectfont
 \setlength{\tabcolsep}{0.3mm}
  \caption{
Performances on low-resource settings by IE models (using T5 Base).
The scores by UIE$^\dag$, which takes additional pre-training with large-scale supervised IE corpus, are copied from raw paper \cite{UIEACL22}.
GEN-5 model takes separate IE modeling on each task, while the other models take unified IE.
}
\begin{center}
\resizebox{1\columnwidth}{!}{
  \begin{tabular}{l|ccccc|cccccc|cccc|c}
\toprule
 \multicolumn{1}{c|}{\multirow{3}{*}{ Task\&Data}}&   \multicolumn{4}{c}{\bf\em Span Extraction} &\phantom{} &  \multicolumn{5}{c}{\bf\em Pair Extraction} &\phantom{}& \multicolumn{3}{c}{\bf\em Hyper-pair Extraction} &\phantom{} & \multicolumn{1}{c}{\multirow{3}{*}{\em Avg.}} \\
\cmidrule(r){2-6}\cmidrule(r){7-12}\cmidrule(r){13-16}
  &   \multicolumn{4}{c}{\scriptsize \bf NER} &\phantom{} & \multicolumn{3}{c}{\scriptsize \bf RE}  & \multicolumn{1}{l}{\scriptsize \bf AOP} & \multicolumn{1}{l}{\scriptsize \bf ASTE} &\phantom{} &  \multicolumn{1}{l}{\scriptsize \bf ORL} & \multicolumn{1}{c}{\scriptsize \bf SRL} & \multicolumn{1}{c}{\scriptsize \bf EE} &\phantom{} & \\
\cmidrule(r){2-6}\cmidrule(r){7-9}\cmidrule(r){10-10}\cmidrule(r){11-11}\cmidrule(r){13-13}\cmidrule(r){14-14}\cmidrule(r){15-15} 
  &     {\scriptsize CoNLL03} & {\scriptsize OntoNote} &  {\scriptsize ACE04} & {\scriptsize ACE05} &\phantom{} &   {\scriptsize CoNLL04} &   {\scriptsize NYT} &   {\scriptsize ACE05} &   {\scriptsize Res14} &   {\scriptsize Res14} &\phantom{} &   {\scriptsize MPQA} &    {\scriptsize CoNLL12} &   {\scriptsize ACE05} &\phantom{} &\\
\midrule

\multicolumn{15}{l}{$\bullet$ \bf\em  1-shot} \\

\quad UIE$^\dag$ &  \bf 46.4 &  / & / & / &\phantom{} &22.1 & / & / &   / & /  &\phantom{}& / & / & /   &\phantom{}& / \\
\quad GEN-T5\emph{+Dep\&ConSyn} & 27.2 &  20.4 &  14.8 &  17.6 &\phantom{} &  8.2 & 25.7 &  10.8 & 12.8 &  10.8 &\phantom{} &  1.1 & 6.5 & 1.5 &\phantom{} & 13.1 \\
\quad UIE\emph{+Dep\&ConSyn} & 30.3 &  23.6 &  17.5 &  20.7 &\phantom{} &  12.8 &  26.7 &  14.3 & 16.7 &  13.0 &\phantom{} &  2.8 & 14.0 &  3.8 &\phantom{} & 16.4 \\
\quad \textbf{LasUIE} & 39.4 &  \bf 47.6 &  \bf 38.5 &  \bf 44.7 &\phantom{} &  \bf 25.7 &  \bf 45.0 &  \bf 26.7 &  \bf 30.0 &  \bf 38.4 &\phantom{} &  \bf 18.9 &  \bf 32.8 &  \bf 23.7  &\phantom{}& \bf 34.3 \\

\cdashline{1-17}
\multicolumn{15}{l}{$\bullet$ \bf\em  10-shot} \\

\quad UIE$^\dag$ &  73.9 &  / & / & / &\phantom{}  & 52.4 & / & / &  / & / &\phantom{} & / & / & / &\phantom{} &/  \\
\quad GEN-T5\emph{+Dep\&ConSyn} & 67.4 &  64.7 &  49.2 &  52.8 &\phantom{} &  45.6 &  50.8 &  37.4 &  19.7 &  17.8 &\phantom{} &  5.4 & 18.7 &  12.2 &\phantom{} &  36.8  \\
\quad UIE\emph{+Dep\&ConSyn} & 69.5 &  68.4 &  52.8 &  54.1 &\phantom{} &  51.8 &  56.0 &  43.8 &  22.5 &  26.1 &\phantom{} &  10.5 &  23.2 &  17.6 &\phantom{} &41.4 \\
\quad \textbf{LasUIE} & \bf 74.0 &  \bf 78.3 &  \bf 60.3 &  \bf 65.3  &\phantom{}&  \bf 55.0 &  \bf 67.7 &  \bf 46.1 &  \bf 42.4 &  \bf 48.8  &\phantom{}&  \bf 25.4 &  \bf 45.8 &  \bf 27.1 &\phantom{} &\bf 53.0 \\

\midrule
\multicolumn{15}{l}{$\bullet$ \bf\em 1\% data} \\

\quad UIE$^\dag$ &  \bf 82.8 &  / & / & /  &\phantom{}& 30.8 & / & / &  / & / &\phantom{} & / & / & / &\phantom{} &/  \\
\quad GEN-T5\emph{+Dep\&ConSyn} & 79.5 &  72.4 &  58.3 &  61.7 &\phantom{} &  17.8 &  35.8 &  15.4 &  15.3 &  15.3 &\phantom{} &  3.3 & 10.7 &  3.4 &\phantom{} & 32.4  \\
\quad UIE\emph{+Dep\&ConSyn} & 80.6 &  73.2 &  60.4 &  63.8 &\phantom{} &  23.5 &  40.4 &  22.7 &  20.6 &  18.5 &\phantom{} &  5.3  & 17.6 & 10.2 &\phantom{} &36.4 \\
\quad \textbf{LasUIE} & 82.1 &  \bf 84.5 &  \bf 65.7 &  \bf 70.1 &\phantom{} &  \bf 32.0 &  \bf 53.6 &  \bf 34.2 &  \bf 34.8 &  \bf 41.7 &\phantom{} &  \bf 21.0 &  \bf 39.8 &  \bf 25.7  &\phantom{}&  \bf 48.8 \\

\cdashline{1-17}

\multicolumn{15}{l}{$\bullet$ \bf\em 10\% data} \\

\quad UIE$^\dag$ &  89.6 &  / & / & / &\phantom{} & 59.2 &  / & / &  / & / &\phantom{} & / & / & / &\phantom{} &/  \\
\quad GEN-T5\emph{+Dep\&ConSyn} & 89.0 &  84.0 &  71.3 &  68.8 &\phantom{} &  52.4 &  80.4 &  45.7 &  56.0 &  59.7 &\phantom{} &  22.4 &  50.7 &  26.7 &\phantom{} &  58.9  \\
\quad UIE\emph{+Dep\&ConSyn} & 89.3 &  85.8 &  72.1 &  70.6 &\phantom{} &  54.9 &  82.5 &  47.6 &  58.3 &  62.6 &\phantom{} &  27.4 &  54.3 &  31.7 &\phantom{} &64.4 \\
\quad \textbf{LasUIE} & \bf 91.6 &  \bf 89.3 &  \bf 83.6 &  \bf 81.7 &\phantom{} &  \bf 60.8 &  \bf 86.0 &  \bf 50.5 &  \bf 63.0 &  \bf 66.7 &\phantom{} &  \bf 36.0 &  \bf 58.4 &  \bf 38.4 &\phantom{} &  \bf 67.2 \\

\bottomrule
\end{tabular}
}
\vspace{-8pt}
\end{center}
\label{main-2}
\end{table*}

\begin{figure}[!t]
\centering
\includegraphics[width=1.0\columnwidth]{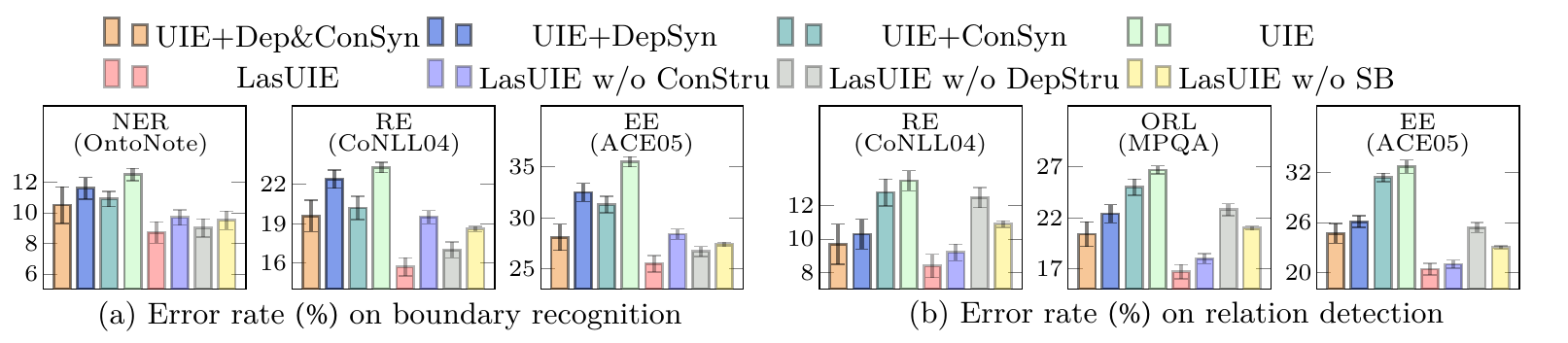}
\caption{
Error rates on boundary recognition and relation detection, respectively.
}
\label{error}
\vspace{-1.4em}
\end{figure}

\paragraph{Q2: {What are the differences to integrate the constituency and dependency syntactic structure?}}
We now observe the results of different tasks in both Table \ref{main-1}\&\ref{main-2}, and we can find that \emph{on the span extraction type IE (i.e., NER) the improvements from constituency syntax prevail, while the dependency type of structure features dominate the pair-wise tasks, i.e., (hyper-)pair extraction.}
We further analyze the error rate on the predictions of two kernel elements of IE, i.e., boundary recognition and relation detection, respectively on various tasks.
We see from Fig. \ref{error} that \emph{the constituency structure more tends to offer key clues for the boundary recognition; 
while the dependent trees are more apt to cope with the relation detection, solving long-range dependence issue.}
This shows that two heterogeneous structures have complementary advantages to UIE.
\emph{Thus, when combining both of them together, all the end tasks receive the enhancements to the greatest extent.}

\paragraph{Q3: {For UIE, is it more advanced for GLM to automatically learn latent structures than injecting external syntax parse trees?}}
First, the comparisons in the main results directly prove the advance of using latent structural features for UIE.
For example, under the fair comparison, our LasUIE beats the UIE\emph{+Dep\&ConSyn} model with average 1.3\%(=75.9-74.6) F1 improvement.
Even comparing with the UIE$^{*\dag}$ that takes additional pre-training on large-scale supervised IE corpus, LasUIE keeps its superiority in almost all cases.
This evidently verifies that \emph{it is necessary for LMs to automatically learn latent structure information for better UIE.}
The underlying reason of our model's improvements could be that \emph{the dynamically learned richer structural knowledge in LasUIE largely avoids the noises that are introduced in external syntax parse annotations.}
Besides, as shown in Fig. \ref{error}, LasUIE reduces the errors on predicting the mention boundaries and relational pairings more significantly than the baseline counterparts.

Further we step into our LasUIE system itself, and inspect the ablation models, M14-M15, as shown in Table \ref{main-1}.
We see that the proposed structural broadcaster module plays important role to the overall system, i.e., without SB, LasUIE is downgraded to the level of UIE\emph{+Dep\&ConSyn}.
Also the structure diversification regularization mechanism serves positive effect.

\vspace{-8pt}
\paragraph{Q4: {Is it necessary to further fine-tune the structures in GLM for UIE?}}

According to the results of the ablation model, M16, in Table \ref{main-1}, we can directly claim the answer is positive.
Without performing structural fine-tuning, the results by LasUIE hurt clearly, with averaged 1.0\%(=75.9-74.9) F1 drop.
We next dig into the structural fine-tuning mechanism, analyzing how the auto-induced structural features influence the UIE performances.

\begin{figure}[!t]
\begin{minipage}[t]{0.49\textwidth}
\includegraphics[width=1\linewidth]{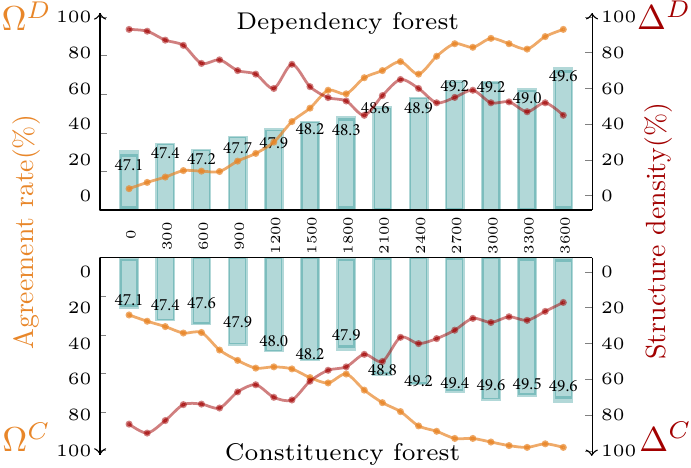}
\caption{\label{FtStru}
\normalsize Trajectories of the changing structure agreement rates and densities during task-oriented structure fine-tuning, based on event extraction (ACE05).
$X$-axis is the iteration steps for fine-tuning.
Bars means the task performances (F1).
}
\end{minipage}
\hfill
\begin{minipage}[t]{0.49\textwidth}
\includegraphics[width=1\linewidth]{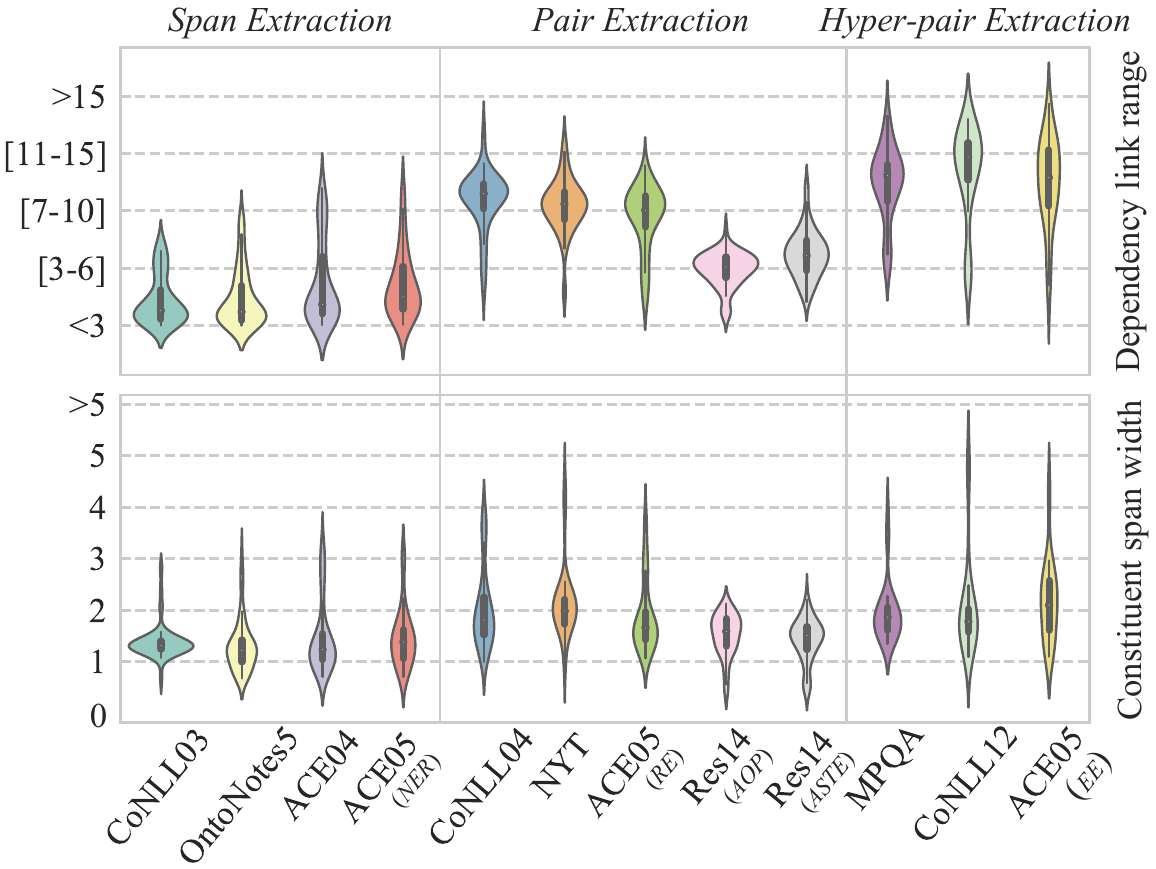}
\caption{\label{structure-distribution}
\normalsize The distributions of the range of word-word dependency link (words) in forest $\mathcal{F}^D$ and the constituency phrasal span width (words) in forest $\mathcal{F}^C$ on each data.
}
\end{minipage}
\vspace{-1.3em}
\end{figure}

We first study the changing trajectories of the 1) the structure agreement rates $\Omega^{D/C}$ and 2) the structure densities $\Delta^{D/C}$ of the structural forests $\mathcal{F}^{D/C}$, during fine-tuning.
$\Omega^{C}$ or $\Omega^{D}$ is defined as the percentage that gold spans correspond to the phrasal spans in the constituency forest $\mathcal{F}^{C}$, or the gold relational pairs coincide with the word-word edges in the dependency forest $\mathcal{F}^{D}$.
As plotted in Fig. \ref{FtStru}, along with the fine-tuning process the task performance climbs gradually.
Meanwhile, both the agreement rate $\Omega^{D/C}$ increases, which means that \emph{the structural fine-tuning indeed can effectively adjust the learned structural information towards task-specific.}
Also, the structural densities of two forests change from dense to sparse, which depicts a structure pruning process in our system.

Finally, in Fig. \ref{structure-distribution} we present the distribution of the range of word-word dependency links in $\mathcal{F}^{D}$ and distribution of the constituency phrasal span width in $\mathcal{F}^{C}$.
We can discover that different end tasks rely on subtly varying structural features or attributes.
For example, hyper-pair extraction tasks require longer-range dependency features for relation determination, compared to the IE tasks of other prototypes.
In turn, this certifies that \emph{our system can correctly learn the peculiar structural bias for a specific IE task}, thanks to the task-oriented structure fine-tuning mechanism.

\vspace{-7pt}
\section{Conclusion and Discussion}

\vspace{-7pt}
This work investigates a novel structure-aware generative language model (GLM) that learns rich heterogeneous syntactic structure representations for better unified information extraction (UIE).
First, a well pre-trained GLM is taken as backbone to reach the goal of UIE, feeding with label prompt-based texts and predicting linearized hierarchical expressions that describe the actual IE target.
During post-training, the proposed heterogeneous structure inductor automatically generates rich structure information without relying on any additional syntax annotation.
A structural broadcaster then compacts various trees into forests for enhancing the structural feature utility and guiding better context generation.
The learned structural knowledge is further fine-tuned on the in-house training data so as to adapt into the task-specific need.
Extensive experiments and in-depth analyses demonstrate the efficacy of our system on improving the UIE.

\textbf{Potential impact and limitations of the work.}
The proposed structure-aware GLM learns syntactic knowledge relies only on the plain texts with easy access, without any cost of large-scale human-labor annotations.
The system will benefit the development of IE community, i.e., training one single unified model for effectively solving various IE tasks, which especially addresses the issue of IE data annotation scarcity in the real-life applications.
One biggest potential risk is that the GPU-based training of our language model will cost  energy consumption and CO$_2$ emissions.

\begin{ack}
This research is supported by the Sea-NExT Joint Lab.
We would also like to thank the anonymous reviewers for their valuable feedbacks.
\end{ack}

\newpage

{\small
\bibliographystyle{plainnat}
\bibliography{ref}

\begin{thebibliography}{88}
\providecommand{\natexlab}[1]{#1}
\providecommand{\url}[1]{\texttt{#1}}
\expandafter\ifx\csname urlstyle\endcsname\relax
  \providecommand{\doi}[1]{doi: #1}\else
  \providecommand{\doi}{doi: \begingroup \urlstyle{rm}\Url}\fi

\bibitem[Agic(2012)]{agic-2012-k}
Zeljko Agic.
\newblock {K}-best spanning tree dependency parsing with verb valency lexicon
  reranking.
\newblock In \emph{Proceedings of the International Conference on Computational
  Linguistics}, pages 1--12, 2012.

\bibitem[Ahmed et~al.(2019)Ahmed, Samee, and Mercer]{ahmed-etal-2019-need}
Mahtab Ahmed, Muhammad~Rifayat Samee, and Robert~E. Mercer.
\newblock You only need attention to traverse trees.
\newblock In \emph{Proceedings of the Annual Meeting of the Association for
  Computational Linguistics}, pages 316--322, 2019.

\bibitem[Alt et~al.(2019)Alt, H{\"u}bner, and Hennig]{alt-etal-2019-fine}
Christoph Alt, Marc H{\"u}bner, and Leonhard Hennig.
\newblock Fine-tuning pre-trained transformer language models to distantly
  supervised relation extraction.
\newblock In \emph{Proceedings of the 57th Annual Meeting of the Association
  for Computational Linguistics}, pages 1388--1398, 2019.

\bibitem[Andersen et~al.(1992)Andersen, Hayes, Weinstein, Huettner, Schmandt,
  and Nirenburg]{andersen-etal-1992-automatic}
Peggy~M. Andersen, Philip~J. Hayes, Steven~P. Weinstein, Alison~K. Huettner,
  Linda~M. Schmandt, and Irene~B. Nirenburg.
\newblock Automatic extraction of facts from press releases to generate news
  stories.
\newblock In \emph{Third Conference on Applied Natural Language Processing},
  pages 170--177, 1992.

\bibitem[Angeli et~al.(2015)Angeli, Johnson~Premkumar, and
  Manning]{angeli-etal-2015-leveraging}
Gabor Angeli, Melvin~Jose Johnson~Premkumar, and Christopher~D. Manning.
\newblock Leveraging linguistic structure for open domain information
  extraction.
\newblock In \emph{Proceedings of the 53rd Annual Meeting of the Association
  for Computational Linguistics and the 7th International Joint Conference on
  Natural Language Processing}, pages 344--354, 2015.

\bibitem[Bai et~al.(2021)Bai, Wang, Chen, Yang, Bai, Yu, and
  Tong]{bai-etal-2021-syntax}
Jiangang Bai, Yujing Wang, Yiren Chen, Yaming Yang, Jing Bai, Jing Yu, and
  Yunhai Tong.
\newblock Syntax-{BERT}: Improving pre-trained transformers with syntax trees.
\newblock In \emph{Proceedings of the 16th Conference of the European Chapter
  of the Association for Computational Linguistics: Main Volume}, pages
  3011--3020, 2021.

\bibitem[Choi et~al.(2018)Choi, Yoo, and Lee]{ChoiYL18}
Jihun Choi, Kang~Min Yoo, and Sang{-}goo Lee.
\newblock Learning to compose task-specific tree structures.
\newblock In \emph{Proceedings of the Thirty-Second {AAAI} Conference on
  Artificial Intelligence}, pages 5094--5101, 2018.

\bibitem[Conneau et~al.(2018)Conneau, Kruszewski, Lample, Barrault, and
  Baroni]{Probing2018}
Alexis Conneau, German Kruszewski, Guillaume Lample, Lo{\"\i}c Barrault, and
  Marco Baroni.
\newblock What you can cram into a single {\$}{\&}!{\#}* vector: Probing
  sentence embeddings for linguistic properties.
\newblock In \emph{Proceedings of the Annual Meeting of the Association for
  Computational Linguistics}, pages 2126--2136, 2018.

\bibitem[Cucerzan and Yarowsky(1999)]{cucerzan-yarowsky-1999-language}
Silviu Cucerzan and David Yarowsky.
\newblock Language independent named entity recognition combining morphological
  and contextual evidence.
\newblock In \emph{Proceedings of the Joint {SIGDAT} Conference on Empirical
  Methods in Natural Language Processing and Very Large Corpora}, 1999.

\bibitem[Devlin et~al.(2019)Devlin, Chang, Lee, and Toutanova]{devlin2019bert}
Jacob Devlin, Ming-Wei Chang, Kenton Lee, and Kristina Toutanova.
\newblock {BERT}: Pre-training of deep bidirectional transformers for language
  understanding.
\newblock In \emph{Proceedings of the 2019 Conference of the North {A}merican
  Chapter of the Association for Computational Linguistics: Human Language
  Technologies}, pages 4171--4186, 2019.

\bibitem[Doddington et~al.(2004)Doddington, Mitchell, Przybocki, Ramshaw,
  Strassel, and Weischedel]{DoddingtonMPRSW04}
George~R. Doddington, Alexis Mitchell, Mark~A. Przybocki, Lance~A. Ramshaw,
  Stephanie~M. Strassel, and Ralph~M. Weischedel.
\newblock The automatic content extraction {(ACE)} program - tasks, data, and
  evaluation.
\newblock In \emph{Proceedings of The Language Resources and Evaluation
  Conference}, page 837–840, 2004.

\bibitem[Drozdov et~al.(2019)Drozdov, Verga, Yadav, Iyyer, and
  McCallum]{drozdov-etal-2019-unsupervised-latent}
Andrew Drozdov, Patrick Verga, Mohit Yadav, Mohit Iyyer, and Andrew McCallum.
\newblock Unsupervised latent tree induction with deep inside-outside recursive
  auto-encoders.
\newblock In \emph{Proceedings of the 2019 Conference of the North {A}merican
  Chapter of the Association for Computational Linguistics: Human Language
  Technologies}, pages 1129--1141, 2019.

\bibitem[Fei et~al.(2020{\natexlab{a}})Fei, Ren, and
  Ji]{fei-etal-2020-retrofitting}
Hao Fei, Yafeng Ren, and Donghong Ji.
\newblock Retrofitting structure-aware transformer language model for end
  tasks.
\newblock In \emph{Proceedings of the 2020 Conference on Empirical Methods in
  Natural Language Processing}, pages 2151--2161, 2020{\natexlab{a}}.

\bibitem[Fei et~al.(2020{\natexlab{b}})Fei, Ren, and Ji]{fei2020boundaries}
Hao Fei, Yafeng Ren, and Donghong Ji.
\newblock Boundaries and edges rethinking: An end-to-end neural model for
  overlapping entity relation extraction.
\newblock \emph{Information Processing \& Management}, 57\penalty0
  (6):\penalty0 102311, 2020{\natexlab{b}}.

\bibitem[Fei et~al.(2021{\natexlab{a}})Fei, Ji, Li, Liu, Ren, and
  Li]{0001JLLRL21}
Hao Fei, Donghong Ji, Bobo Li, Yijiang Liu, Yafeng Ren, and Fei Li.
\newblock Rethinking boundaries: End-to-end recognition of discontinuous
  mentions with pointer networks.
\newblock In \emph{Proceedings of the AAAI Conference on Artificial
  Intelligence}, pages 12785--12793, 2021{\natexlab{a}}.

\bibitem[Fei et~al.(2021{\natexlab{b}})Fei, Ren, Zhang, Ji, and
  Liang]{FeiRZJL21}
Hao Fei, Yafeng Ren, Yue Zhang, Donghong Ji, and Xiaohui Liang.
\newblock Enriching contextualized language model from knowledge graph for
  biomedical information extraction.
\newblock \emph{Briefings in Bioinformatics}, 22\penalty0 (3),
  2021{\natexlab{b}}.

\bibitem[Fei et~al.(2021{\natexlab{c}})Fei, Wu, Ren, Li, and
  Ji]{fei-etal-2021-better}
Hao Fei, Shengqiong Wu, Yafeng Ren, Fei Li, and Donghong Ji.
\newblock Better combine them together! integrating syntactic constituency and
  dependency representations for semantic role labeling.
\newblock In \emph{Findings of the Association for Computational Linguistics:
  ACL-IJCNLP 2021}, pages 549--559, 2021{\natexlab{c}}.

\bibitem[Fei et~al.(2022)Fei, Wu, Ren, and Zhang]{0001WRZ22}
Hao Fei, Shengqiong Wu, Yafeng Ren, and Meishan Zhang.
\newblock Matching structure for dual learning.
\newblock In \emph{Proceedings of the International Conference on Machine
  Learning, {ICML}}, pages 6373--6391, 2022.

\bibitem[Gildea and Jurafsky(2000)]{gildea-jurafsky-2000-automatic}
Daniel Gildea and Daniel Jurafsky.
\newblock Automatic labeling of semantic roles.
\newblock In \emph{Proceedings of the Annual Meeting of the Association for
  Computational Linguistics}, pages 512--520, 2000.

\bibitem[Goldberg(2019)]{Assessing190105287}
Yoav Goldberg.
\newblock Assessing bert's syntactic abilities.
\newblock \emph{CoRR}, abs/1901.05287, 2019.

\bibitem[Halpin and Moore(2006)]{halpin-moore-2006-event}
Harry Halpin and Johanna~D. Moore.
\newblock Event extraction in a plot advice agent.
\newblock In \emph{Proceedings of the 21st International Conference on
  Computational Linguistics and 44th Annual Meeting of the Association for
  Computational Linguistics}, pages 857--864, 2006.

\bibitem[hristopher Walker et~al.(2006)hristopher Walker, Strassel, Medero, and
  Maeda]{ACE5}
hristopher Walker, Stephanie Strassel, Julie Medero, and Kazuaki Maeda.
\newblock Ace 2005 multilingual training corpus.
\newblock In \emph{Proceedings of Philadelphia: Linguistic Data Consortium},
  2006.

\bibitem[Jawahar et~al.(2019)Jawahar, Sagot, and Seddah]{jawahar2019}
Ganesh Jawahar, Beno{\^\i}t Sagot, and Djam{\'e} Seddah.
\newblock What does {BERT} learn about the structure of language?
\newblock In \emph{Proceedings of the Annual Meeting of the Association for
  Computational Linguistics}, pages 3651--3657, 2019.

\bibitem[Jie and Lu(2019)]{jie-lu-2019-dependency}
Zhanming Jie and Wei Lu.
\newblock Dependency-guided {LSTM}-{CRF} for named entity recognition.
\newblock In \emph{Proceedings of the 2019 Conference on Empirical Methods in
  Natural Language Processing and the 9th International Joint Conference on
  Natural Language Processing}, pages 3862--3872, 2019.

\bibitem[Joshi et~al.(2020)Joshi, Chen, Liu, Weld, Zettlemoyer, and
  Levy]{JoshiCLWZL20}
Mandar Joshi, Danqi Chen, Yinhan Liu, Daniel~S. Weld, Luke Zettlemoyer, and
  Omer Levy.
\newblock Spanbert: Improving pre-training by representing and predicting
  spans.
\newblock \emph{Transactions of the Association for Computational Linguistics},
  8:\penalty0 64--77, 2020.

\bibitem[Kawintiranon and Singh(2021)]{kawintiranon-singh-2021-knowledge}
Kornraphop Kawintiranon and Lisa Singh.
\newblock Knowledge enhanced masked language model for stance detection.
\newblock In \emph{Proceedings of the 2021 Conference of the North American
  Chapter of the Association for Computational Linguistics: Human Language
  Technologies}, pages 4725--4735, 2021.

\bibitem[Kim and Hovy(2006)]{kim2006extracting}
Soo-Min Kim and Eduard Hovy.
\newblock Extracting opinions, opinion holders, and topics expressed in online
  news media text.
\newblock In \emph{Proceedings of the Workshop on Sentiment and Subjectivity in
  Text}, pages 1--8, 2006.

\bibitem[Kim et~al.(2019)Kim, Dyer, and Rush]{kim-etal-2019-compound}
Yoon Kim, Chris Dyer, and Alexander Rush.
\newblock Compound probabilistic context-free grammars for grammar induction.
\newblock In \emph{Proceedings of the Annual Meeting of the Association for
  Computational Linguistics}, pages 2369--2385, 2019.

\bibitem[Kolluru et~al.(2020)Kolluru, Aggarwal, Rathore, {Mausam}, and
  Chakrabarti]{kolluru-etal-2020-imojie}
Keshav Kolluru, Samarth Aggarwal, Vipul Rathore, {Mausam}, and Soumen
  Chakrabarti.
\newblock {IM}o{JIE}: Iterative memory-based joint open information extraction.
\newblock In \emph{Proceedings of the 58th Annual Meeting of the Association
  for Computational Linguistics}, pages 5871--5886, 2020.

\bibitem[Kovaleva et~al.(2019)Kovaleva, Romanov, Rogers, and
  Rumshisky]{kovaleva-etal-2019-revealing}
Olga Kovaleva, Alexey Romanov, Anna Rogers, and Anna Rumshisky.
\newblock Revealing the dark secrets of {BERT}.
\newblock In \emph{Proceedings of the 2019 Conference on Empirical Methods in
  Natural Language Processing and the 9th International Joint Conference on
  Natural Language Processing}, pages 4365--4374, 2019.

\bibitem[Lepori et~al.(2020)Lepori, Linzen, and
  McCoy]{lepori-etal-2020-representations}
Michael Lepori, Tal Linzen, and R.~Thomas McCoy.
\newblock Representations of syntax {[MASK]} useful: {E}ffects of constituency
  and dependency structure in recursive {LSTM}s.
\newblock In \emph{Proceedings of the 58th Annual Meeting of the Association
  for Computational Linguistics}, pages 3306--3316, 2020.

\bibitem[Lewis et~al.(2020)Lewis, Liu, Goyal, Ghazvininejad, Mohamed, Levy,
  Stoyanov, and Zettlemoyer]{lewis-etal-2020-bart}
Mike Lewis, Yinhan Liu, Naman Goyal, Marjan Ghazvininejad, Abdelrahman Mohamed,
  Omer Levy, Veselin Stoyanov, and Luke Zettlemoyer.
\newblock {BART}: Denoising sequence-to-sequence pre-training for natural
  language generation, translation, and comprehension.
\newblock In \emph{Proceedings of the Annual Meeting of the Association for
  Computational Linguistics}, pages 7871--7880, 2020.

\bibitem[Li et~al.(2021{\natexlab{a}})Li, Lin, Zhang, and
  Ji]{li-etal-2021-span}
Fei Li, ZhiChao Lin, Meishan Zhang, and Donghong Ji.
\newblock A span-based model for joint overlapped and discontinuous named
  entity recognition.
\newblock In \emph{Proceedings of the 59th Annual Meeting of the Association
  for Computational Linguistics and the 11th International Joint Conference on
  Natural Language Processing}, pages 4814--4828, 2021{\natexlab{a}}.

\bibitem[Li et~al.(2021{\natexlab{b}})Li, Xu, Li, Fei, Ren, and
  Ji]{li-etal-2021-mrn}
Jingye Li, Kang Xu, Fei Li, Hao Fei, Yafeng Ren, and Donghong Ji.
\newblock {MRN}: A locally and globally mention-based reasoning network for
  document-level relation extraction.
\newblock In \emph{Findings of the Association for Computational Linguistics:
  ACL-IJCNLP 2021}, pages 1359--1370, 2021{\natexlab{b}}.

\bibitem[Li et~al.(2022)Li, Fei, Liu, Wu, Zhang, Teng, Ji, and Li]{W2NER22}
Jingye Li, Hao Fei, Jiang Liu, Shengqiong Wu, Meishan Zhang, Chong Teng,
  Donghong Ji, and Fei Li.
\newblock Unified named entity recognition as word-word relation
  classification.
\newblock In \emph{Proceedings of the AAAI Conference on Artificial
  Intelligence}, pages 10965--10973, 2022.

\bibitem[Li et~al.(2020)Li, Feng, Meng, Han, Wu, and Li]{li-etal-2020-unified}
Xiaoya Li, Jingrong Feng, Yuxian Meng, Qinghong Han, Fei Wu, and Jiwei Li.
\newblock A unified {MRC} framework for named entity recognition.
\newblock In \emph{Proceedings of the 58th Annual Meeting of the Association
  for Computational Linguistics}, pages 5849--5859, 2020.

\bibitem[Li and Lam(2017)]{li-lam-2017-deep}
Xin Li and Wai Lam.
\newblock Deep multi-task learning for aspect term extraction with memory
  interaction.
\newblock In \emph{Proceedings of the Conference on Empirical Methods in
  Natural Language Processing}, pages 2886--2892, 2017.

\bibitem[Li et~al.(2021{\natexlab{c}})Li, Zhou, Li, Xu, and
  Cao]{li-etal-2021-improving-bert}
Zhongli Li, Qingyu Zhou, Chao Li, Ke~Xu, and Yunbo Cao.
\newblock Improving {BERT} with syntax-aware local attention.
\newblock In \emph{Findings of the Association for Computational Linguistics:
  ACL-IJCNLP 2021}, pages 645--653, 2021{\natexlab{c}}.

\bibitem[Lin et~al.(2018)Lin, Lu, Han, and Sun]{lin-etal-2018-adaptive}
Hongyu Lin, Yaojie Lu, Xianpei Han, and Le~Sun.
\newblock Adaptive scaling for sparse detection in information extraction.
\newblock In \emph{Proceedings of the 56th Annual Meeting of the Association
  for Computational Linguistics}, pages 1033--1043, July 2018.

\bibitem[Lin et~al.(2020)Lin, Ji, Huang, and Wu]{lin-etal-2020-joint}
Ying Lin, Heng Ji, Fei Huang, and Lingfei Wu.
\newblock A joint neural model for information extraction with global features.
\newblock In \emph{Proceedings of the 58th Annual Meeting of the Association
  for Computational Linguistics}, pages 7999--8009, 2020.

\bibitem[Liu et~al.(2019)Liu, Du, and
  Stoyanov]{liu-etal-2019-knowledge-augmented}
Angli Liu, Jingfei Du, and Veselin Stoyanov.
\newblock Knowledge-augmented language model and its application to
  unsupervised named-entity recognition.
\newblock In \emph{Proceedings of the 2019 Conference of the North {A}merican
  Chapter of the Association for Computational Linguistics: Human Language
  Technologies}, pages 1142--1150, 2019.

\bibitem[Lu et~al.(2022)Lu, Liu, Dai, Xiao, Lin, Han, Sun, and Wu]{UIEACL22}
Yaojie Lu, Qing Liu, Dai Dai, Xinyan Xiao, Hongyu Lin, Xianpei Han, Le~Sun, and
  Hua Wu.
\newblock Unified structure generation for universal information extraction.
\newblock In \emph{Proceedings of the 60th Annual Meeting of the Association
  for Computational Linguistics}, pages 5755--5772, 2022.

\bibitem[Ma et~al.(2018)Ma, Tamura, Utiyama, Zhao, and
  Sumita]{ma-etal-2018-forest}
Chunpeng Ma, Akihiro Tamura, Masao Utiyama, Tiejun Zhao, and Eiichiro Sumita.
\newblock Forest-based neural machine translation.
\newblock In \emph{Proceedings of the 56th Annual Meeting of the Association
  for Computational Linguistics}, pages 1253--1263, 2018.

\bibitem[Ma and Hovy(2016)]{ma-hovy-2016-end}
Xuezhe Ma and Eduard Hovy.
\newblock End-to-end sequence labeling via bi-directional {LSTM}-{CNN}s-{CRF}.
\newblock In \emph{Proceedings of the 54th Annual Meeting of the Association
  for Computational Linguistics}, pages 1064--1074, 2016.

\bibitem[Marcheggiani and Titov(2017)]{marcheggiani-titov-2017-encoding}
Diego Marcheggiani and Ivan Titov.
\newblock Encoding sentences with graph convolutional networks for semantic
  role labeling.
\newblock In \emph{Proceedings of the Conference on Empirical Methods in
  Natural Language Processing}, pages 1506--1515, 2017.

\bibitem[Marcheggiani and Titov(2020)]{marcheggiani-titov-2020-graph}
Diego Marcheggiani and Ivan Titov.
\newblock Graph convolutions over constituent trees for syntax-aware semantic
  role labeling.
\newblock In \emph{Proceedings of the 2020 Conference on Empirical Methods in
  Natural Language Processing}, pages 3915--3928, 2020.

\bibitem[Maslennikov and Chua(2007)]{maslennikov-chua-2007-multi}
Mstislav Maslennikov and Tat-Seng Chua.
\newblock A multi-resolution framework for information extraction from free
  text.
\newblock In \emph{Proceedings of the 45th Annual Meeting of the Association of
  Computational Linguistics}, pages 592--599, 2007.

\bibitem[Mi and Huang(2008)]{mi-huang-2008-forest}
Haitao Mi and Liang Huang.
\newblock Forest-based translation rule extraction.
\newblock In \emph{Proceedings of the 2008 Conference on Empirical Methods in
  Natural Language Processing}, pages 206--214, 2008.

\bibitem[Miwa and Bansal(2016)]{miwa-bansal-2016-end}
Makoto Miwa and Mohit Bansal.
\newblock End-to-end relation extraction using {LSTM}s on sequences and tree
  structures.
\newblock In \emph{Proceedings of the 54th Annual Meeting of the Association
  for Computational Linguistics}, pages 1105--1116, 2016.

\bibitem[Peng et~al.(2020)Peng, Xu, Bing, Huang, Lu, and Si]{PengXBHLS20}
Haiyun Peng, Lu~Xu, Lidong Bing, Fei Huang, Wei Lu, and Luo Si.
\newblock Knowing what, how and why: {A} near complete solution for
  aspect-based sentiment analysis.
\newblock In \emph{Proceedings of the AAAI Conference on Artificial
  Intelligence}, pages 8600--8607, 2020.

\bibitem[Pontiki et~al.(2014)Pontiki, Galanis, Pavlopoulos, Papageorgiou,
  Androutsopoulos, and Manandhar]{pontiki-etal-2014-semeval}
Maria Pontiki, Dimitris Galanis, John Pavlopoulos, Harris Papageorgiou, Ion
  Androutsopoulos, and Suresh Manandhar.
\newblock {S}em{E}val-2014 task 4: Aspect based sentiment analysis.
\newblock In \emph{Proceedings of the 8th International Workshop on Semantic
  Evaluation ({S}em{E}val 2014)}, pages 27--35, 2014.

\bibitem[Pouran Ben~Veyseh et~al.(2020)Pouran Ben~Veyseh, Nguyen, and
  Nguyen]{pouran-ben-veyseh-etal-2020-graph}
Amir Pouran Ben~Veyseh, Tuan~Ngo Nguyen, and Thien~Huu Nguyen.
\newblock Graph transformer networks with syntactic and semantic structures for
  event argument extraction.
\newblock In \emph{Findings of the Association for Computational Linguistics:
  EMNLP 2020}, pages 3651--3661, 2020.

\bibitem[Pradhan et~al.(2013)Pradhan, Moschitti, Xue, Ng, Bj{\"o}rkelund,
  Uryupina, Zhang, and Zhong]{pradhan-etal-2013-towards}
Sameer Pradhan, Alessandro Moschitti, Nianwen Xue, Hwee~Tou Ng, Anders
  Bj{\"o}rkelund, Olga Uryupina, Yuchen Zhang, and Zhi Zhong.
\newblock Towards robust linguistic analysis using {O}nto{N}otes.
\newblock In \emph{Proceedings of the Seventeenth Conference on Computational
  Natural Language Learning}, pages 143--152, 2013.

\bibitem[Quan et~al.(2022)Quan, Min, Li, and Yang]{quan2022compound}
Xiaojun Quan, Zhengcheng Min, Kun Li, and Yunyi Yang.
\newblock Compound aspect extraction by augmentation and constituency lattice.
\newblock \emph{IEEE Transactions on Affective Computing}, 2022.

\bibitem[Raffel et~al.(2020)Raffel, Shazeer, Roberts, Lee, Narang, Matena,
  Zhou, Li, and Liu]{RaffelSRLNMZLL20}
Colin Raffel, Noam Shazeer, Adam Roberts, Katherine Lee, Sharan Narang, Michael
  Matena, Yanqi Zhou, Wei Li, and Peter~J. Liu.
\newblock Exploring the limits of transfer learning with a unified text-to-text
  transformer.
\newblock \emph{Journal of Machine Learning Research}, 21:\penalty0
  140:1--140:67, 2020.

\bibitem[Riedel et~al.(2010)Riedel, Yao, and McCallum]{RiedelYM10}
Sebastian Riedel, Limin Yao, and Andrew McCallum.
\newblock Modeling relations and their mentions without labeled text.
\newblock In \emph{Proceedings of Machine Learning and Knowledge Discovery in
  Databases, European Conference, {ECML} {PKDD}}, pages 148--163, 2010.

\bibitem[Roth and Yih(2004)]{roth-yih-2004-linear}
Dan Roth and Wen-tau Yih.
\newblock A linear programming formulation for global inference in natural
  language tasks.
\newblock In \emph{Proceedings of the Eighth Conference on Computational
  Natural Language Learning ({C}o{NLL}-2004) at {HLT}-{NAACL}}, pages 1--8,
  2004.

\bibitem[Shen et~al.(2018)Shen, Lin, Huang, and Courville]{ShenLHC18}
Yikang Shen, Zhouhan Lin, Chin{-}Wei Huang, and Aaron~C. Courville.
\newblock Neural language modeling by jointly learning syntax and lexicon.
\newblock In \emph{Proceedings of the International Conference on Learning
  Representations}, 2018.

\bibitem[Shen et~al.(2019)Shen, Tan, Sordoni, and Courville]{ShenTSC19}
Yikang Shen, Shawn Tan, Alessandro Sordoni, and Aaron~C. Courville.
\newblock Ordered neurons: Integrating tree structures into recurrent neural
  networks.
\newblock In \emph{Proceedings of the International Conference on Learning
  Representations}, 2019.

\bibitem[Shen et~al.(2021)Shen, Tay, Zheng, Bahri, Metzler, and
  Courville]{shen-etal-2021-structformer}
Yikang Shen, Yi~Tay, Che Zheng, Dara Bahri, Donald Metzler, and Aaron
  Courville.
\newblock {S}truct{F}ormer: Joint unsupervised induction of dependency and
  constituency structure from masked language modeling.
\newblock In \emph{Proceedings of the 59th Annual Meeting of the Association
  for Computational Linguistics and the 11th International Joint Conference on
  Natural Language Processing}, pages 7196--7209, 2021.

\bibitem[Shi et~al.(2022)Shi, Li, Li, Fei, and Ji]{shi-etal-2022-effective}
Wenxuan Shi, Fei Li, Jingye Li, Hao Fei, and Donghong Ji.
\newblock Effective token graph modeling using a novel labeling strategy for
  structured sentiment analysis.
\newblock In \emph{Proceedings of the 60th Annual Meeting of the Association
  for Computational Linguistics}, pages 4232--4241, 2022.

\bibitem[Song et~al.(2019)Song, Zhang, Gildea, Yu, Wang, and
  Su]{song-etal-2019-leveraging}
Linfeng Song, Yue Zhang, Daniel Gildea, Mo~Yu, Zhiguo Wang, and Jinsong Su.
\newblock Leveraging dependency forest for neural medical relation extraction.
\newblock In \emph{Proceedings of the 2019 Conference on Empirical Methods in
  Natural Language Processing and the 9th International Joint Conference on
  Natural Language Processing}, pages 208--218, 2019.

\bibitem[Surdeanu et~al.(2003)Surdeanu, Harabagiu, Williams, and
  Aarseth]{surdeanu-etal-2003-using}
Mihai Surdeanu, Sanda Harabagiu, John Williams, and Paul Aarseth.
\newblock Using predicate-argument structures for information extraction.
\newblock In \emph{Proceedings of the 41st Annual Meeting of the Association
  for Computational Linguistics}, pages 8--15, 2003.

\bibitem[Tang et~al.(2016)Tang, Qin, Feng, and Liu]{tang-etal-2016-effective}
Duyu Tang, Bing Qin, Xiaocheng Feng, and Ting Liu.
\newblock Effective {LSTM}s for target-dependent sentiment classification.
\newblock In \emph{Proceedings of the International Conference on Computational
  Linguistics}, pages 3298--3307, 2016.

\bibitem[Tenney et~al.(2019)Tenney, Xia, Chen, Wang, Poliak, McCoy, Kim, Durme,
  Bowman, Das, and Pavlick]{Probing19}
Ian Tenney, Patrick Xia, Berlin Chen, Alex Wang, Adam Poliak, R.~Thomas McCoy,
  Najoung Kim, Benjamin~Van Durme, Samuel~R. Bowman, Dipanjan Das, and Ellie
  Pavlick.
\newblock What do you learn from context? probing for sentence structure in
  contextualized word representations.
\newblock In \emph{Proceedings of the International Conference on Learning
  Representations}, 2019.

\bibitem[Tjong Kim~Sang and
  De~Meulder(2003)]{tjong-kim-sang-de-meulder-2003-introduction}
Erik~F. Tjong Kim~Sang and Fien De~Meulder.
\newblock Introduction to the {C}o{NLL}-2003 shared task: Language-independent
  named entity recognition.
\newblock In \emph{Proceedings of the Seventh Conference on Natural Language
  Learning at {HLT}-{NAACL}}, pages 142--147, 2003.

\bibitem[Velickovic et~al.(2018)Velickovic, Cucurull, Casanova, Romero,
  Li{\`{o}}, and Bengio]{VelickovicCCRLB18}
Petar Velickovic, Guillem Cucurull, Arantxa Casanova, Adriana Romero, Pietro
  Li{\`{o}}, and Yoshua Bengio.
\newblock Graph attention networks.
\newblock In \emph{Proceedings of the International Conference on Learning
  Representations}, 2018.

\bibitem[Vig and Belinkov(2019)]{vig-belinkov-2019-analyzing}
Jesse Vig and Yonatan Belinkov.
\newblock Analyzing the structure of attention in a transformer language model.
\newblock In \emph{Proceedings of the 2019 ACL Workshop BlackboxNLP: Analyzing
  and Interpreting Neural Networks for NLP}, pages 63--76, 2019.

\bibitem[Wang and Lu(2020)]{wang-lu-2020-two}
Jue Wang and Wei Lu.
\newblock Two are better than one: Joint entity and relation extraction with
  table-sequence encoders.
\newblock In \emph{Proceedings of the 2020 Conference on Empirical Methods in
  Natural Language Processing}, pages 1706--1721, 2020.

\bibitem[Wang et~al.(2021{\natexlab{a}})Wang, Jiang, Bach, Wang, Huang, Huang,
  and Tu]{wang-etal-2021-improving}
Xinyu Wang, Yong Jiang, Nguyen Bach, Tao Wang, Zhongqiang Huang, Fei Huang, and
  Kewei Tu.
\newblock Improving named entity recognition by external context retrieving and
  cooperative learning.
\newblock In \emph{Proceedings of the 59th Annual Meeting of the Association
  for Computational Linguistics and the 11th International Joint Conference on
  Natural Language Processing}, pages 1800--1812, 2021{\natexlab{a}}.

\bibitem[Wang et~al.(2019)Wang, Lee, and Chen]{wang-etal-2019-tree}
Yaushian Wang, Hung-Yi Lee, and Yun-Nung Chen.
\newblock Tree transformer: Integrating tree structures into self-attention.
\newblock In \emph{Proceedings of the Conference on Empirical Methods in
  Natural Language Processing}, pages 1061--1070, 2019.

\bibitem[Wang et~al.(2020)Wang, Yu, Zhang, Liu, Zhu, and Sun]{wang2020tplinker}
Yucheng Wang, Bowen Yu, Yueyang Zhang, Tingwen Liu, Hongsong Zhu, and Limin
  Sun.
\newblock Tplinker: Single-stage joint extraction of entities and relations
  through token pair linking.
\newblock In \emph{Proceedings of the International Conference on Computational
  Linguistics}, pages 1572--1582, 2020.

\bibitem[Wang et~al.(2021{\natexlab{b}})Wang, Wang, Han, Lin, Hou, Liu, Li, Li,
  and Zhou]{wang-etal-2021-cleve}
Ziqi Wang, Xiaozhi Wang, Xu~Han, Yankai Lin, Lei Hou, Zhiyuan Liu, Peng Li,
  Juanzi Li, and Jie Zhou.
\newblock {CLEVE}: {C}ontrastive {P}re-training for {E}vent {E}xtraction.
\newblock In \emph{Proceedings of the 59th Annual Meeting of the Association
  for Computational Linguistics and the 11th International Joint Conference on
  Natural Language Processing}, pages 6283--6297, 2021{\natexlab{b}}.

\bibitem[Wiebe et~al.(2005)Wiebe, Wilson, and Cardie]{WiebeWC05}
Janyce Wiebe, Theresa Wilson, and Claire Cardie.
\newblock Annotating expressions of opinions and emotions in language.
\newblock \emph{Language Resources and Evaluation}, 39\penalty0 (2-3):\penalty0
  165--210, 2005.

\bibitem[Williams et~al.(2018)Williams, Drozdov, and Bowman]{WilliamsDB18}
Adina Williams, Andrew Drozdov, and Samuel~R. Bowman.
\newblock Do latent tree learning models identify meaningful structure in
  sentences?
\newblock \emph{Transactions of the Association for Computational Linguistics},
  6:\penalty0 253--267, 2018.

\bibitem[Williams(1992)]{Williams92}
Ronald~J. Williams.
\newblock Simple statistical gradient-following algorithms for connectionist
  reinforcement learning.
\newblock \emph{Machine Learning}, 8:\penalty0 229--256, 1992.

\bibitem[Wu et~al.(2021)Wu, Fei, Ren, Ji, and Li]{Wu0RJL21}
Shengqiong Wu, Hao Fei, Yafeng Ren, Donghong Ji, and Jingye Li.
\newblock Learn from syntax: Improving pair-wise aspect and opinion terms
  extraction with rich syntactic knowledge.
\newblock In \emph{Proceedings of the Thirtieth International Joint Conference
  on Artificial Intelligence}, pages 3957--3963, 2021.

\bibitem[Wu et~al.(2022)Wu, Fei, Li, Ji, Zhang, Liu, and Teng]{ORL22}
Shengqiong Wu, Hao Fei, Fei Li, Donghong Ji, Meishan Zhang, Yijiang Liu, and
  Chong Teng.
\newblock Mastering the explicit opinion-role interaction: Syntax-aided neural
  transition system for unified opinion role labeling.
\newblock In \emph{Proceedings of the AAAI Conference on Artificial
  Intelligence}, pages 3943--3951, 2022.

\bibitem[Xu et~al.(2015)Xu, Ringlstetter, Kim, Kondrak, Goebel, and
  Miyao]{xu-etal-2015-lexicalized}
Ying Xu, Christoph Ringlstetter, Mi-Young Kim, Grzegorz Kondrak, Randy Goebel,
  and Yusuke Miyao.
\newblock A lexicalized tree kernel for open information extraction.
\newblock In \emph{Proceedings of the 53rd Annual Meeting of the Association
  for Computational Linguistics and the 7th International Joint Conference on
  Natural Language Processing}, pages 279--284, 2015.

\bibitem[Yan et~al.(2021)Yan, Gui, Dai, Guo, Zhang, and
  Qiu]{yan-etal-2021-unified-generative}
Hang Yan, Tao Gui, Junqi Dai, Qipeng Guo, Zheng Zhang, and Xipeng Qiu.
\newblock A unified generative framework for various {NER} subtasks.
\newblock In \emph{Proceedings of the 59th Annual Meeting of the Association
  for Computational Linguistics and the 11th International Joint Conference on
  Natural Language Processing}, pages 5808--5822, 2021.

\bibitem[Zelenko et~al.(2002)Zelenko, Aone, and
  Richardella]{zelenko-etal-2002-kernel}
Dmitry Zelenko, Chinatsu Aone, and Anthony Richardella.
\newblock Kernel methods for relation extraction.
\newblock In \emph{Proceedings of the Conference on Empirical Methods in
  Natural Language Processing}, pages 71--78, 2002.

\bibitem[Zeng et~al.(2018)Zeng, Zeng, He, Liu, and
  Zhao]{zeng-etal-2018-extracting}
Xiangrong Zeng, Daojian Zeng, Shizhu He, Kang Liu, and Jun Zhao.
\newblock Extracting relational facts by an end-to-end neural model with copy
  mechanism.
\newblock In \emph{Proceedings of the 56th Annual Meeting of the Association
  for Computational Linguistics}, pages 506--514, 2018.

\bibitem[Zhang et~al.(2021{\natexlab{a}})Zhang, Cai, Wang, Qiu, Yang, and
  He]{zhang-etal-2021-smedbert}
Taolin Zhang, Zerui Cai, Chengyu Wang, Minghui Qiu, Bite Yang, and Xiaofeng He.
\newblock {SM}ed{BERT}: A knowledge-enhanced pre-trained language model with
  structured semantics for medical text mining.
\newblock In \emph{Proceedings of the 59th Annual Meeting of the Association
  for Computational Linguistics and the 11th International Joint Conference on
  Natural Language Processing}, pages 5882--5893, 2021{\natexlab{a}}.

\bibitem[Zhang et~al.(2021{\natexlab{b}})Zhang, Li, Deng, Bing, and
  Lam]{zhang-etal-2021-towards-generative}
Wenxuan Zhang, Xin Li, Yang Deng, Lidong Bing, and Wai Lam.
\newblock Towards generative aspect-based sentiment analysis.
\newblock In \emph{Proceedings of the 59th Annual Meeting of the Association
  for Computational Linguistics and the 11th International Joint Conference on
  Natural Language Processing}, pages 504--510, 2021{\natexlab{b}}.

\bibitem[Zhao et~al.(2020)Zhao, Huang, Zhang, Lu, and Xue]{ZhaoHZLX20}
He~Zhao, Longtao Huang, Rong Zhang, Quan Lu, and Hui Xue.
\newblock Spanmlt: {A} span-based multi-task learning framework for pair-wise
  aspect and opinion terms extraction.
\newblock In \emph{Proceedings of the Annual Meeting of the Association for
  Computational Linguistics}, pages 3239--3248, 2020.

\bibitem[Zheng et~al.(2021)Zheng, Wen, Chen, Yang, Zhang, Zhang, Zhang, Qin,
  Ming, and Zheng]{zheng-etal-2021-prgc}
Hengyi Zheng, Rui Wen, Xi~Chen, Yifan Yang, Yunyan Zhang, Ziheng Zhang, Ningyu
  Zhang, Bin Qin, Xu~Ming, and Yefeng Zheng.
\newblock {PRGC}: Potential relation and global correspondence based joint
  relational triple extraction.
\newblock In \emph{Proceedings of the 59th Annual Meeting of the Association
  for Computational Linguistics and the 11th International Joint Conference on
  Natural Language Processing}, pages 6225--6235, 2021.

\bibitem[Zhong and Chen(2021)]{zhong-chen-2021-frustratingly}
Zexuan Zhong and Danqi Chen.
\newblock A frustratingly easy approach for entity and relation extraction.
\newblock In \emph{Proceedings of the 2021 Conference of the North American
  Chapter of the Association for Computational Linguistics: Human Language
  Technologies}, pages 50--61, 2021.

\bibitem[Zmigrod et~al.(2021)Zmigrod, Vieira, and
  Cotterell]{zmigrod-etal-2021-finding}
Ran Zmigrod, Tim Vieira, and Ryan Cotterell.
\newblock On finding the k-best non-projective dependency trees.
\newblock In \emph{Proceedings of the 59th Annual Meeting of the Association
  for Computational Linguistics and the 11th International Joint Conference on
  Natural Language Processing}, pages 1324--1337, 2021.

\end{thebibliography}
}


\section*{Checklist}

\begin{enumerate}

\item For all authors...
\begin{enumerate}
  \item Do the main claims made in the abstract and introduction accurately reflect the paper's contributions and scope?
    \answerYes{}
  \item Did you describe the limitations of your work?
    \answerYes{}
  \item Did you discuss any potential negative societal impacts of your work?
    \answerNA{}
  \item Have you read the ethics review guidelines and ensured that your paper conforms to them?
    \answerYes{}
\end{enumerate}

\item If you are including theoretical results...
\begin{enumerate}
  \item Did you state the full set of assumptions of all theoretical results?
    \answerYes{}
  \item Did you include complete proofs of all theoretical results?
    \answerNA{}
\end{enumerate}

\item If you ran experiments...
\begin{enumerate}
  \item Did you include the code, data, and instructions needed to reproduce the main experimental results (either in the supplemental material or as a URL)?
    \answerYes{}
  \item Did you specify all the training details (e.g., data splits, hyperparameters, how they were chosen)?
    \answerYes{}
  \item Did you report error bars (e.g., with respect to the random seed after running experiments multiple times)?
    \answerYes{}
  \item Did you include the total amount of compute and the type of resources used (e.g., type of GPUs, internal cluster, or cloud provider)?
    \answerYes{}
\end{enumerate}

\item If you are using existing assets (e.g., code, data, models) or curating/releasing new assets...
\begin{enumerate}
  \item If your work uses existing assets, did you cite the creators?
    \answerNA{}
  \item Did you mention the license of the assets?
    \answerNA{}
  \item Did you include any new assets either in the supplemental material or as a URL?
    \answerNA{}
  \item Did you discuss whether and how consent was obtained from people whose data you're using/curating?
    \answerYes{}
  \item Did you discuss whether the data you are using/curating contains personally identifiable information or offensive content?
    \answerNA{}
\end{enumerate}

\item If you used crowdsourcing or conducted research with human subjects...
\begin{enumerate}
  \item Did you include the full text of instructions given to participants and screenshots, if applicable?
    \answerNA{}
  \item Did you describe any potential participant risks, with links to Institutional Review Board (IRB) approvals, if applicable?
    \answerNA{}
  \item Did you include the estimated hourly wage paid to participants and the total amount spent on participant compensation?
    \answerNA{}
\end{enumerate}

\end{enumerate}

\newpage

\appendix

\section{Detail of Syntax Structure Induction}
\label{Detail of Syntax Structure Induction}

Here we unfold the induction of Eq. (\ref{eqn1}) for the structure induction.
Based on the rule $\Gamma_C$ we have:
\begin{equation} \nonumber
\setlength\abovedisplayskip{2pt}
\setlength\belowdisplayskip{2pt}
p(o^d_i>o^c_k) = \sigma(o^d_i>o^c_k) \,.
\end{equation}
Then, the probability that the $l$-th ($l < i$) token is inside $C_{[l,r]}$ is equal to the probability that $o^d_i$ is larger then the maximum $o^c_k$ between $l$ and $i$:
\begin{equation} \nonumber
\setlength\abovedisplayskip{2pt}
\setlength\belowdisplayskip{2pt}
p(l \in C_{[l,r]}) = p(o^d_i> \text{Max}(o^c_{i-1},\cdots,o^c_l)) = \sigma(o^d_i - \mathop{\text{Max}}_{ k\in [l,i)}(o^c_k)  ) \,.
\end{equation}
Thus, the conditional probabilistic of $p(w_l|w_i)$ becomes:
\begin{equation} \nonumber
\setlength\abovedisplayskip{2pt}
\setlength\belowdisplayskip{2pt}
p(w_l|w_i) =\sigma(o^d_i - \mathop{\text{Max}}_{ k\in [l,i)}(o^c_k)  ) -  \sigma(o^d_i - \mathop{\text{Max}}_{k\in (l,i)}(o^c_k)  ) \,.
\end{equation}
Similarly, for $p(w_r|w_i)$:
\begin{equation} \nonumber
\setlength\abovedisplayskip{2pt}
\setlength\belowdisplayskip{2pt}
p(w_r|w_i) = \sigma(o^d_i - \mathop{\text{Max}}_{k\in [i,r)}(o^c_k)  ) -  \sigma(o^d_i - \mathop{\text{Max}}_{k\in [i,r]}(o^c_k)  ) \,.
\end{equation}
And finally, we can derive the probability of phrasal span $C_{[l,r]}$:
\begin{equation} \nonumber
\setlength\abovedisplayskip{2pt}
\setlength\belowdisplayskip{2pt}
\begin{aligned} \nonumber
p_c(c_k|w_i) &= p(w_l|w_i)\cdot p(w_r|w_i) \\
     &= [\sigma(o^d_i - \mathop{\text{Max}}_{ k\in [l,i)}(o^c_k)  ) -  \sigma(o^d_i - \mathop{\text{Max}}_{k\in (l,i)}(o^c_k)  )] \cdot
     [\sigma(o^d_i - \mathop{\text{Max}}_{k\in [i,r)}(o^c_k)  ) -  \sigma(o^d_i - \mathop{\text{Max}}_{k\in [i,r]}(o^c_k)  )] \,.
\end{aligned}
\end{equation}

\section{Baseline Specification}
\label{Baseline specification}

In our experiments we take the current SoTA methods as the separate IE comparisons on each specific task and data.
Here Table \ref{baselines} shows the specifications of these baseline models.
Parts of the results are directly copied from their raw papers, where the Large version LM or GLM is used, while part of the results are from our reimplementation.

\begin{table}[!h]
\fontsize{9}{13}\selectfont
 \setlength{\tabcolsep}{4mm}
  \caption{
SoTA baseline systems for different IE tasks.
}
\begin{center}
  \begin{tabular}{clccc}
\toprule
\bf Task & \bf  Dataset & \bf   Model & \bf   LM Type & \bf   Result Source \\
\midrule
\multirow{3}{*}{NER} & CoNLL03   & \citet{wang-etal-2021-improving} &   RoBERTa-Large &   Raw paper \\
 &  ACE04/05 &  \citet{yan-etal-2021-unified-generative} &  BART-Large &  Raw paper \\
 &  OntoNote &   \citet{W2NER22} &   BERT-Large &  Reimplementation \\

\cdashline{1-5}
\multirow{3}{*}{RE} &   CoNLL04 &   \citet{wang-lu-2020-two} &  ALBERT-large &  Raw paper \\
 &    NYT &   \citet{zheng-etal-2021-prgc} &  BERT-Large &  Raw paper \\
 &    ACE05 &   \citet{zhong-chen-2021-frustratingly} &   ALBERT-XXLarge &  Raw paper \\
\cdashline{1-5}
AOP  & Res14 &  \citet{Wu0RJL21} &  BERT-Large &  Reimplementation \\
\cdashline{1-5}
ASTE &  Res14 &   \citet{zhang-etal-2021-towards-generative} &  BERT-Large &  Reimplementation \\
\cdashline{1-5}
ORL &   MPQA &  \citet{ORL22} &   BERT-Large &  Reimplementation \\
\cdashline{1-5}
SRL  & CoNLL12 &  \citet{fei-etal-2021-better} &  RoBERTa-Large &   Reimplementation \\
\cdashline{1-5}
EE &  ACE05 &   \citet{lin-etal-2020-joint} &   BERT-Large &  Reimplementation \\

\bottomrule
  \end{tabular}
\end{center}
  \label{baselines}
\end{table}

\section{Task Label Prompts}
\label{Task label prompts}

Each task with each dataset will come with different task labels, including the \emph{span attribute labels} and \emph{relation type labels}, which will be used as the label prompts in input.
Table \ref{labels-prompts} summarizes all the label prompts of each dataset.
Also we note that, instead of directly taking the raw label abbreviations as label prompts, we use the full names of labels in natural languages, such that we can fully utilize their semantic representations in GLM.

\newcommand{\tabincell}[2]{\begin{tabular}{@{}#1@{}}#2\end{tabular}}

\begin{table}[!t]
\fontsize{9}{11.3}\selectfont
  \caption{
  The span attribute labels and relation type labels of different tasks and datasets for building the task labels prompts.
  We note that we replace the raw label abbreviations with their full names in natural languages, so as to fully utilize their semantic representations in GLM.
}
\begin{center}
\resizebox{1.0\columnwidth}{!}{
  \begin{tabular}{ccll}
\toprule
\bf Task & \bf  Dataset &    \multicolumn{1}{c}{\bf Span attribute labels} & \multicolumn{1}{c}{\bf Relation type labels} \\
\midrule
\multirow{10}{*}{NER} & CoNLL03 &	\tabincell{l}{location, organization, person,\\ miscellaneous} &	\multicolumn{1}{c}{/} \\
\cdashline{2-4}
 & OntoNote & 	\tabincell{l}{person, nationality, facility, organization,\\ geographical political, location, product,\\ event, work of art, law, language, time, \\date, percent, money, quantity, \\ordinal, cardinal} & 	\multicolumn{1}{c}{/}  \\
\cdashline{2-4}
 & ACE04 & 	\tabincell{l}{person, organization, location, facility,\\ geographical political, vehicle, weapon} & 	\multicolumn{1}{c}{/}  \\
\cdashline{2-4}
 & ACE05 & 	\tabincell{l}{person, organization, location, facility,\\ geographical political, vehicle, weapon} & 	\multicolumn{1}{c}{/}  \\

\hline
\multirow{15}{*}{RE} &   CoNLL04 & 	location, organization, people, other & 	\tabincell{l}{kill, live in, located in,\\ organization in, work for}\\
\cdashline{2-4}
 & NYT & 	location, organization, person	 & 	\tabincell{l}{administrative divisions, advisors,\\ capital, children, company,\\ contains, country, ethnicity,\\ founders, geographic, distribution,\\ industry, location, major shareholder\\ of, major shareholders, nationality,\\ neighborhood of, people,\\ place founded, place lived,\\ place of birth, place of\\ death, profession, religion, teams} \\
\cdashline{2-4}
 & ACE05	 & 	\tabincell{l}{person, organization, location, facility,\\ geographical political, vehicle, weapon}	 & 	\tabincell{l}{agent artifact, general affiliation,\\ organization affiliation, part whole,\\ personal social, physical} \\

\hline
AOP  & Res14 &  aspect, opinion &	default relation \\

\hline
ASTE &  Res14 &  aspect, opinion &	positive, neutral, negative \\

\hline
ORL &   MPQA &  opinion, role &	holder, target \\

\hline
SRL  & CoNLL12 & default argument, default predicate &  \tabincell{l}{agent (ARG0), patient (ARG1),\\ instrument and end state (ARG2),\\ starting point and benefactive and\\ attribute (ARG3), ending point (ARG4),\\ direction (DIR), location (LOC),\\ manner (MNR), extent (EXT),\\ reciprocals (REC), secondary\\ predication (PRD), purpose (PNC),\\ cause (CAU), discourse and\\ connectives (DIS), adverbial and\\ general purpose (ADV), modal verb\\ (MOD), negation (NEG), time (TMP)} \\

\hline
EE &  ACE05 &   \tabincell{l}{default argument, acquit, appeal,\\ arrest jail, attack, born, charge indict,\\ convict, declare bankruptcy, demonstrate,\\ die, divorce, elect, end organization, end,\\ position, execute, extradite, fine, injure,\\ marry, meet, merge organization, nominate,\\ pardon, phone write, release parole,\\ sentence, start organization, start position,\\ sue, transfer money, transfer, ownership,\\ transport, trial hearing} &	\tabincell{l}{person, agent, victim, instrument,\\ attacker, target, instrument, time,\\ place, artifact, vehicle, price, origin,\\ destination, time, buyer, seller,\\ beneficiary, giver, recipient, beneficiary,\\ org, money, entity, position, crime,\\ defendant, prosecutor, adjudicator,\\ plaintiff, sentence} \\

\bottomrule
  \end{tabular}
}
\end{center}
  \label{labels-prompts}
\end{table}

\end{document}